\documentclass[10pt,twocolumn,letterpaper]{article}
\usepackage{algorithm}
\usepackage{algorithmic}
\usepackage{tikz}
\usetikzlibrary{positioning, shapes, arrows.meta}
\usepackage{bm}
\usepackage[table]{xcolor}
\usepackage{graphicx}
\usepackage{makecell}
\usepackage{amsmath}
\usepackage{float}
\usepackage{hhline}
\usepackage{geometry}
\usepackage{booktabs}
\usepackage{arydshln}
\usepackage{amsfonts}
\usepackage{multirow}
\usepackage{subcaption}
\usepackage[table]{xcolor}  

\usepackage{cvpr}              

\usepackage{tikz}

\definecolor{deepred}{RGB}{200,0,0}
\definecolor{deeporange}{RGB}{230,120,20}
\definecolor{deepgold}{RGB}{184,134,11}
\definecolor{deepgreen}{RGB}{0,128,0}
\definecolor{deepblue}{RGB}{0,0,180}
\definecolor{deepviolet}{RGB}{138,43,226}

\definecolor{cvprblue}{rgb}{0.21,0.49,0.74}

\usepackage[pagebackref,breaklinks,colorlinks,
            citecolor=cvprblue,
            linkcolor=cvprblue,
            urlcolor=cvprblue]{hyperref}
\renewcommand*{\backref}[1]{}
\renewcommand*{\backrefalt}[4]{%
  \ifcase #1 \or {\color{cvprblue}#2}\else {\color{cvprblue}#2}\fi}
\def\paperID{} 
\def\confName{CVPR}
\def\confYear{2026}

\title{PhaSR: Generalized Image Shadow Removal with Physically Aligned Priors}

\author{Chia-Ming Lee$^{1,2}$ \quad Yu-Fan Lin$^{2}$ \quad Yu-Jou Hsiao$^{2}$ \quad Jin-Hui Jiang$^{1}$ \\\quad Yu-Lun Liu$^1$ \quad Chih-Chung Hsu$^{1,2}$\\
{$^1$National Yang Ming Chiao Tung University} \quad {$^2$National Cheng Kung University}}

\begin{document}


\maketitle

\begin{abstract}
Shadow removal under diverse lighting conditions requires disentangling illumination from intrinsic reflectance—a challenge compounded when physical priors are not properly aligned. 
We propose \textbf{PhaSR} (\textit{Physically Aligned Shadow Removal}), addressing this through dual-level prior alignment to enable robust performance from single-light shadows to multi-source ambient lighting.
First, Physically Aligned Normalization (PAN) performs closed-form illumination correction via Gray-world normalization, log-domain Retinex decomposition, and dynamic range recombination, suppressing chromatic bias.
Second, Geometric-Semantic Rectification Attention (GSRA) extends differential attention to cross-modal alignment, harmonizing depth-derived geometry with DINO-v2 semantic embeddings to resolve modal conflicts under varying illumination.
Experiments show competitive performance in shadow removal with lower complexity and generalization to ambient lighting where traditional methods fail under multi-source illumination. Our source code is available at  \hyperlink{https://github.com/ming053l/PhaSR}{https://github.com/ming053l/PhaSR}.

\end{abstract}
\vspace{-0.5cm}
\section{Introduction}
\label{sec:intro}
Shadows, as natural consequences of light-object interactions, are ubiquitous optical phenomena that profoundly impact multimedia content analysis, degrading performance in tasks ranging from remote sensing \cite{rtcs}, segmentation \cite{10678598}, tracking \cite{5597618}, and 3D reconstruction \cite{Weder_2023_CVPR,bolanos2024gsc} to multimedia applications \cite{9918057}. 
Removing shadows from images is not only a fundamental computer vision task but also critical for enhancing downstream application performance \cite{8378149,murali2016survey,tiwari2016survey}. 
The core challenge lies in accurately distinguishing shadows from intrinsic object darkness and leveraging contextual information to perform physically plausible color correction and content restoration within shadowed areas \cite{le2019shadow,yang2012shadow}.

\begin{figure}[t!]
\centering
\includegraphics[width=1.0\linewidth]{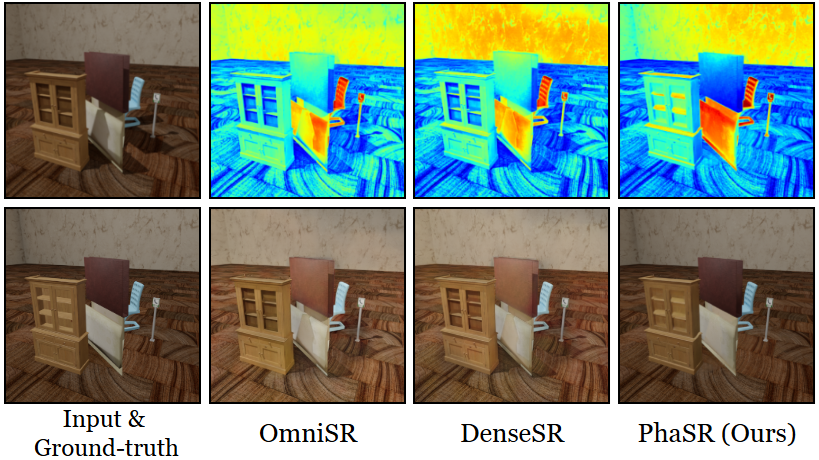}
\vspace{-0.45cm}
\caption{\textbf{Results on indoor-synthesized dataset \cite{omnisr}.}  
Compared with OmniSR \cite{omnisr} and DenseSR \cite{densesr}, PhaSR with the proposed GSRA achieves more accurate boundary localization 
and recovers fine reflectance details even under complex indirect illumination.}
\label{fig:output} 
\vspace{-0.55cm}
\end{figure}

\begin{figure}[t!]
\centering
\includegraphics[width=0.95\linewidth]{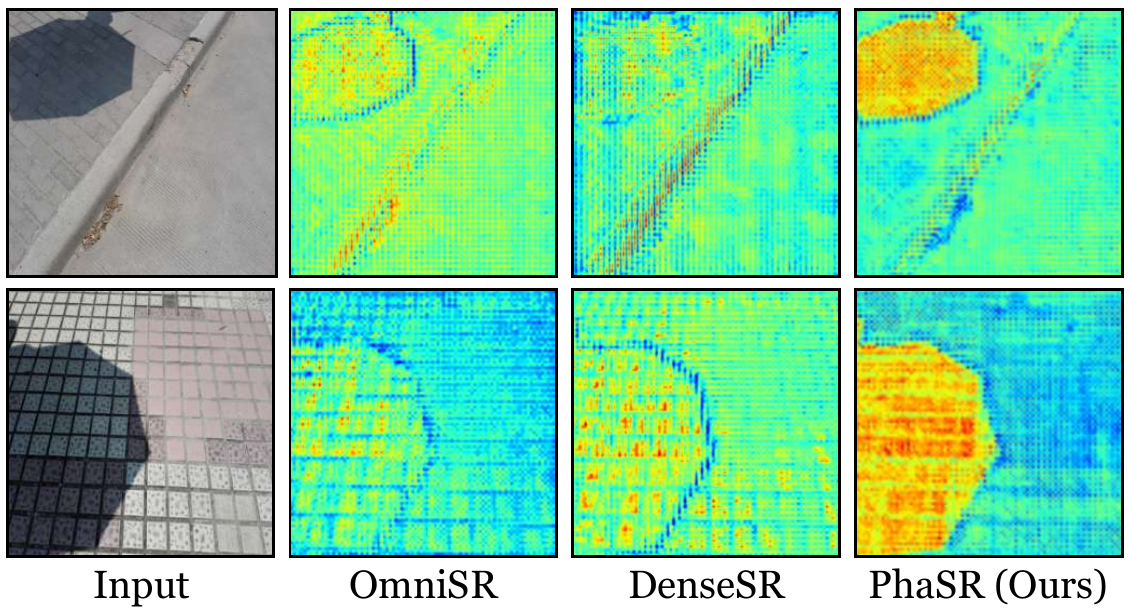}
\caption{\textbf{Intermediate feature visualization.}
Existing methods struggle to leverage physical priors without valid shadow masks under complex environmental lighting. In contrast, our PhaSR precisely highlights and restores shadow regions in both bottleneck and decoder stages, demonstrating strong generalization.
}
\label{fig:featuremap}
\vspace{-0.55cm}
\end{figure}

Despite progress in learning-based shadow removal, key challenges persist. \textbf{First}, shadows are easily confused with intrinsic material properties when relying solely on RGB cues, causing color distortion near textured boundaries. \textbf{Second}, while existing methods achieve strong performance on single-light direct shadow benchmarks, emerging applications demand generalization to more complex scenarios—indoor ambient lighting with multiple sources, color shifts, and diffuse indirect illumination—where prior works show limited robustness (Figure~\ref{fig:output}). \textbf{Third}, conventional encoder-decoder frameworks fail to effectively propagate physical priors, as uniform fusion overlooks spatially varying degradation, resulting in blurred edges. As shown in Figure~\ref{fig:featuremap}, prior methods lose physical priors through the bottleneck, whereas our framework preserves them under complex illumination and details recovery.

These challenges stem from \textit{prior misalignment}. While physical-guided feature transformation~\cite{rehit,Xiao_2024_CVPR,fu2021auto,rrlnet} and explicit geometric-semantic prior integration~\cite{omnisr,densesr,xu2024detailpreservinglatentdiffusionstable,des3} enhance robustness, these priors encode conflicting signals: geometric features respond to local shading variations, while semantic features remain stable across lighting. Without proper alignment, geometric noise disrupts semantic consistency, or semantic over-smoothing erases illumination boundaries—particularly problematic for indirect lighting where geometric and semantic cues must cooperate to disentangle ambient effects from surface properties.

We propose \textbf{PhaSR}—\textit{Physically Aligned Shadow Removal}—addressing prior misalignment through two mechanisms. \textit{Physically Aligned Normalization (PAN)} performs model-free preprocessing via Gray-world normalization, log-domain Retinex decomposition, and dynamic range recombination, suppressing global chromatic bias. \textit{Geometric-Semantic Rectification Attention (GSRA)} extends differential attention~\cite{diff} to cross-modal alignment: computing $\mathbf{A}_{\text{rect}} = \mathbf{A}_{\text{sem}} - \lambda \cdot \mathbf{A}_{\text{geo}}$ across depth-derived geometry (DepthAnything-v2~\cite{depthganythingv2}) and semantic embeddings (DINO-v2~\cite{dinov2}) to harmonize local geometric precision with global semantic stability. This explicit alignment enables accurate interpretation of both direct shadows (geometric-dominant) and ambient lighting (semantic-guided), generalizing from single-light to multi-source illumination scenarios. In summary, the main contributions of this work are threefold:
\begin{itemize}
    \item We introduce Physically Aligned Normalization, a closed-form preprocessing module performing Gray-world normalization, log-domain Retinex decomposition, and dynamic range recombination to suppress chromatic bias. PAN \emph{consistently improves existing architectures} by 0.15–0.34 dB across diverse lighting conditions.

 \item We propose geometric-semantic rectification attention, extending differential attention to cross-modal prior alignment. By explicitly incorporating depth-derived geometry and DINO-v2 semantic embeddings, GSRA harmonizes physically grounded geometric precision with semantic stability, addressing modal misalignment challenges in ambient lighting normalization.
    
    \item We demonstrate state-of-the-art performance on challenging shadow removal benchmarks, achieving robust generalization from outdoor direct shadows to indoor indirect and ambient lighting scenarios while maintaining computational efficiency.
\end{itemize}


\section{Related Work}
\label{sec:related}
\noindent\textbf{Single Image Shadow Removal.}
Single-image shadow removal aims to recover the true appearance beneath shadows.
Early traditional methods followed a two-stage pipeline—shadow detection and removal—based on handcrafted features and physical or statistical illumination models \cite{shor2008shadow,zhang2015shadow,cucchiara2003detecting,salamati2011removing,le2019shadow}, but relied on strong priors that struggled with complex lighting and soft shadows.
Deep learning has significantly advanced the field: CNNs \cite{ronneberger2015u, qu2017deshadownet} capture multi-scale features but face locality limits; Transformer-based methods \cite{shadowformer, Xiao_2024_CVPR, RASM} offer better global context, though some rely on explicit shadow masks \cite{shadowformer,Xiao_2024_CVPR}; Diffusion-based models \cite{guo2023shadowdiffusion, mei2024latent,xu2024detailpreservinglatentdiffusionstable} achieve high quality at significant computational cost. 
DeS3 \cite{des3} pioneered using pretrained DINO \cite{dinov2} priors with diffusion for shadow removal, while RRLNet \cite{rrlnet} applied Retinex decomposition to guide diffusion-based texture refinement \cite{shadowrefiner}. 
OmniSR \cite{omnisr} introduced a synthesized dataset and semantic-geometric aware network for both direct and indirect shadows. 
DenseSR \cite{densesr} reframes shadow removal as dense prediction, leveraging geometric-semantic priors and adaptive fusion to overcome ambiguity and boundary blurring. 
ShadowHack \cite{shadowhack} divides shadow removal into luminance recovery and color restoration using rectified outreach attention.

\noindent\textbf{Ambient Light Normalization.}
Beyond conventional shadow removal, recent work has explored the more challenging task of \emph{ambient light normalization (ALN)}~\cite{ambient6k,cl3an}, which addresses complex real-world scenarios with multiple light sources, color shifts, and diffuse indirect illumination. ReHiT~\cite{rehit} achieves efficient mask-free shadow removal through Retinex-guided dual-branch decomposition, while IFBlend~\cite{ambient6k} and RLN2-Lf~\cite{cl3an} extend this framework to multi-source white and RGB color lighting respectively. However, the ill-posed nature of disentangling multiple overlapping light contributions remains an open challenge. Our method, while primarily designed for shadow removal, demonstrates strong generalization to ALN scenarios through physically aligned normalization that handles diverse illumination conditions.
\begin{figure*}[t!]
\centering
\includegraphics[width=1.01\linewidth]{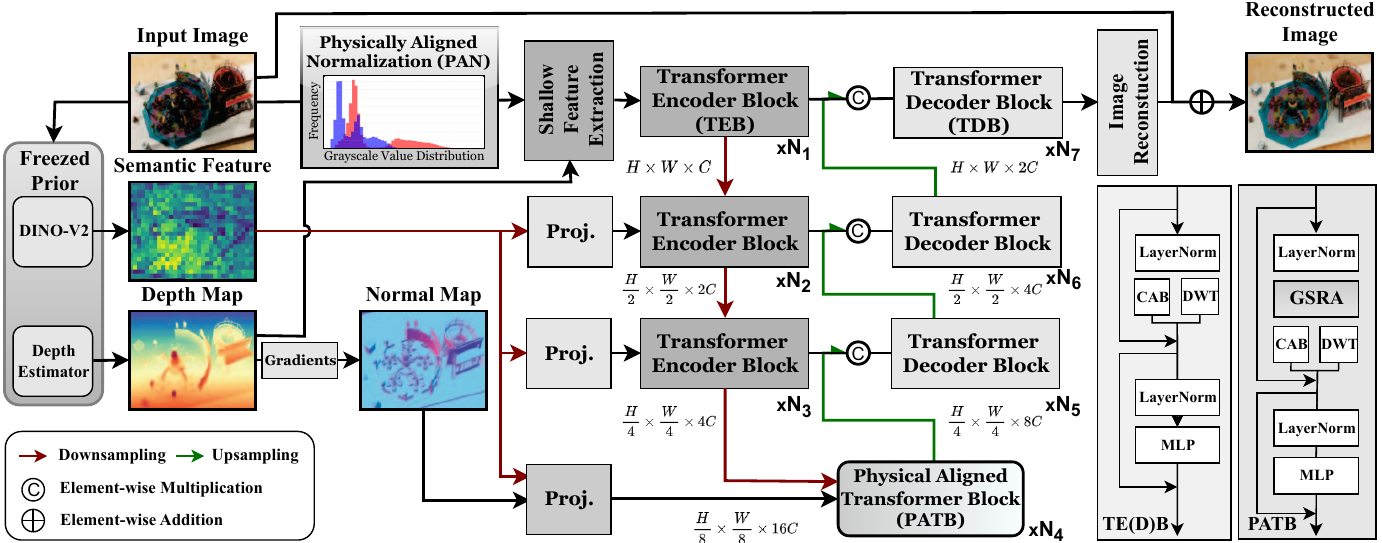}
\vspace{-0.6cm}
\caption{\textbf{Overview of PhaSR: Physically Aligned Shadow Removal.}  
PhaSR achieves physical alignment through two synergistic stages. \textit{Stage 1 (Sec.~\ref{sec:pan})}: PAN performs model-free illumination normalization via Gray-world color correction, log-domain Retinex decomposition ($\log \mathbf{I} = \log \mathbf{R} + \log \mathbf{S}$), and dynamic range recombination, suppressing chromatic bias while preserving reflectance cues. \textit{Stage 2 (Sec.~\ref{sec:gsra})}: The multi-scale Transformer encoder-decoder integrates explicit physical priors—frozen DINO-v2~\cite{dinov2} semantic embeddings at encoder stages and DepthAnything-v2~\cite{depthganythingv2} geometric priors (depth, normals) at the bottleneck—aligned through GSRA's cross-modal differential attention ($\mathbf{A}_{\text{rect}} = \mathbf{A}_{\text{sem}} - \lambda \cdot \mathbf{A}_{\text{geo}}$). This dual-stage physical alignment—global illumination correction followed by local geometric-semantic rectification—enables robust reflectance recovery under complex lighting without requiring shadow masks.}
\label{fig:flowchart}
\vspace{-0.2cm}
\end{figure*}

\noindent\textbf{Intrinsic Decomposition and Physical Priors.}
The formation of shadows originates from the fundamental physics of light transport being occluded by 3D scene geometry \cite{malik2025neural,DiffusionRenderer,unirelight,yang2023sireir}.
Occlusion of direct illumination leads to sharp, well-defined shadows, while indirect illumination—arising from interreflections and ambient scattering—produces soft, graded shadows that are more challenging to model and recover accurately \cite{omnisr}.

From the intrinsic image perspective, the observed image $\mathbf{I}$ can be decomposed into its albedo and shading components:
\begin{equation}
\mathbf{I}(x) = \mathbf{A}(x) \otimes \mathbf{S}(x),
\end{equation}
where $\mathbf{A}$ denotes the intrinsic surface reflectance and $\mathbf{S}$ represents spatially varying illumination.
This formulation, grounded in Retinex theory, provides a physically interpretable basis for shadow removal since shadows mainly alter the shading component without changing the underlying albedo \cite{divideandconquer,shadowhack,rrlnet,rehit}.
Recovering $\mathbf{A}$ and $\mathbf{S}$ from a single shadowed observation, however, is inherently ill-posed due to ambiguity between dark materials and shaded regions, indirect lighting complexity, and spatially non-uniform degradation.
Recent advances leverage large-scale pretrained models as sources of \textbf{physical priors}—including depth and normal maps \cite{depthganythingv2}, semantic features \cite{dinov2,sam}, and illumination cues via Retinex decomposition \cite{LIRM}—to guide more consistent intrinsic decomposition and illumination reasoning \cite{divideandconquer,rrlnet,omnisr,densesr,shadowhack}.
Building on these advances, PhaSR integrates closed-form Retinex decomposition with explicit geometric-semantic prior alignment, enabling robust illumination normalization across shadow removal and ambient lighting scenarios.

\section{Methodology}
\label{sec:methodology}

\textbf{Overview.}
Figure~\ref{fig:flowchart} presents PhaSR, which achieves physically aligned shadow removal through \textit{dual-level prior alignment}. At the \textit{global level}, PAN performs parameter-free illumination-reflectance decomposition via log-domain Retinex theory, aligning the input image with the physical assumption that shadows alter illumination while preserving intrinsic surface properties. This closed-form preprocessing suppresses chromatic bias induced by colored illuminants and stabilizes luminance distribution, providing an illumination-consistent foundation for subsequent reasoning. At the \textit{local level}, GSRA rectifies geometric and semantic priors through cross-modal differential attention, aligning depth-derived shading cues (which respond to local light-geometry interactions) with semantic embeddings (which encode material identity stable across lighting changes). By computing $\mathbf{A}_{\text{rect}} = \mathbf{A}_{\text{sem}} - \lambda \cdot \mathbf{A}_{\text{geo}}$, GSRA resolves the conflicting responses between modalities—suppressing geometric noise in uniformly lit regions while preserving geometric precision at true illumination boundaries. This dual-level alignment—global illumination normalization followed by local geometric-semantic rectification—enables the network to disentangle reflectance from complex lighting effects, generalizing from single-light direct shadows to multi-source ambient illumination without requiring shadow masks.

\begin{figure}[t]
    \centering
    \includegraphics[width=1\linewidth]{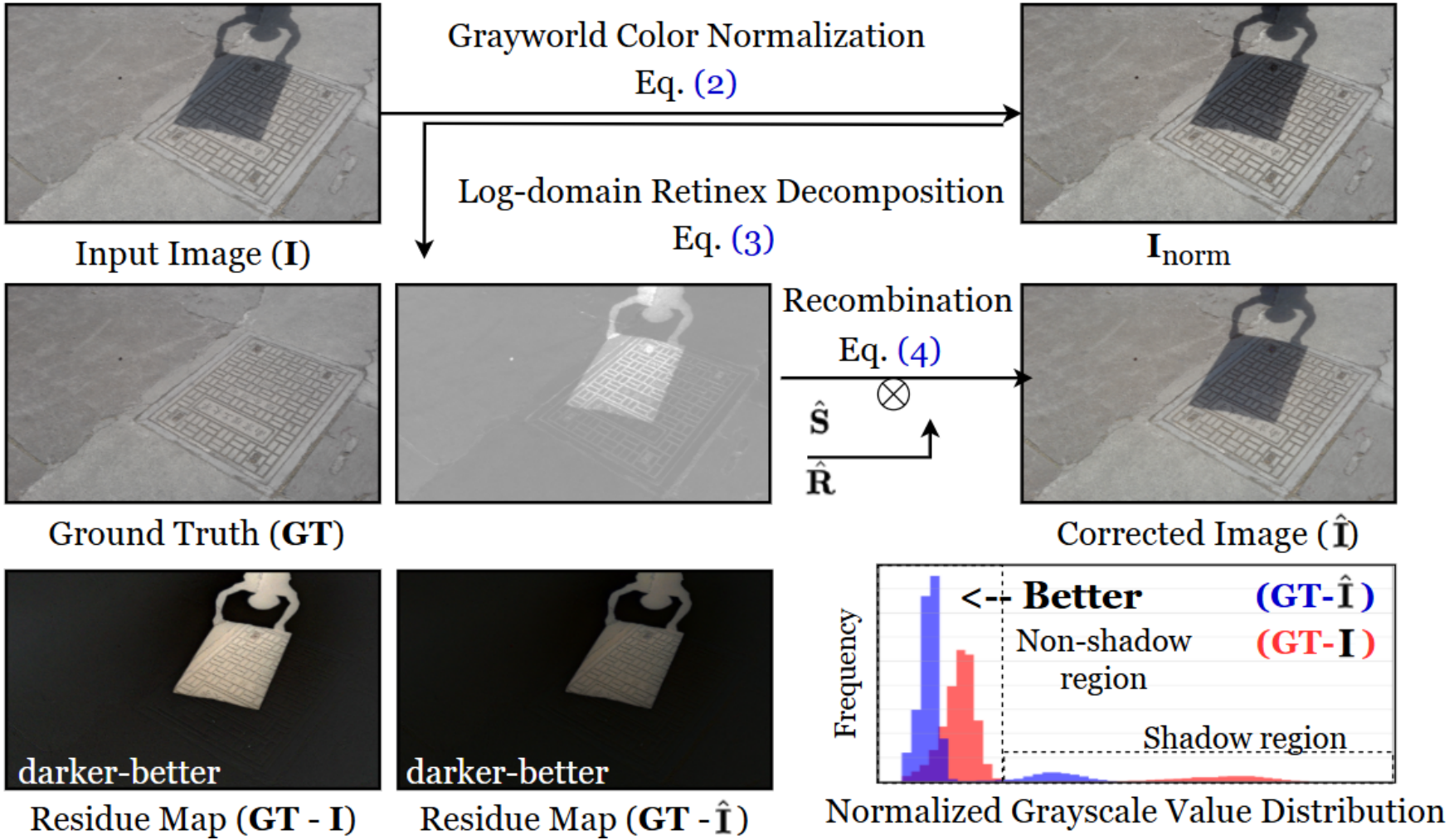}
    \caption{\textbf{Overview of the proposed PAN.} 
    It performs model-free illumination correction through three stages: 
    (1) Global color normalization removes chromatic bias, 
    (2) log-domain Retinex decomposition separates reflectance $\hat{\mathbf{R}}$ from illumination $\hat{\mathbf{S}}$ via closed-form operations, and 
    (3) recombination produces the illumination-consistent output $\hat{\mathbf{I}}$. 
    }
    \label{fig:pan_pipeline}
    \vspace{-0.3cm}
\end{figure}

\subsection{Physically Aligned Normalization}
\label{sec:pan}

Recent advances in Retinex-based shadow removal~\cite{rehit,shadowhack} have demonstrated that explicit illumination-reflectance decomposition provides crucial inductive biases for handling lighting variations. ReHiT~\cite{rehit} employs dual-branch networks for illumination-guided shadow removal, RLN²~\cite{cl3an} extends this to RGB color lighting via HSV-guided decomposition, and IFBlend~\cite{ambient6k} addresses multi-source white lighting through image-frequency joint entropy maximization. Classical methods such as ACE~\cite{ace} and color constancy approaches~\cite{colorconstancy} similarly leverage illumination priors for perceptual uniformity.

Inspired by these works, PAN adopts the Retinex formulation but implements it through closed-form operations rather than learned mappings, achieving illumination invariance via model-free log-domain decomposition that normalizes color statistics while preserving reflectance cues.

\textbf{Gray-world Color Normalization.}  
Real-world images often exhibit chromatic bias induced by illuminant color (e.g., warm indoor lighting, cool daylight), which confounds subsequent reflectance-illumination decomposition. Under the Gray-world assumption~\cite{grayworld}, we first perform:
\begin{equation}
    \mathbf{I}_{\text{norm}} = \mathbf{I} \cdot \frac{\mathbb{E}[\mathbf{I}]}{\mathbb{E}_c[\mathbf{I}] + \varepsilon},
\end{equation}
where $\mathbb{E}[\mathbf{I}]$ denotes spatial average, $\mathbb{E}_c[\cdot]$ represents per-channel mean, and $\varepsilon = 10^{-6}$ prevents division by zero. This balances channel-wise illumination, removing color casts while stabilizing overall luminance for subsequent decomposition.

\textbf{Log-domain Retinex Decomposition.}  
We then disentangle illumination from reflectance following the image formation model:
\begin{equation}
\mathbf{I}_{\text{norm}}(x) = \mathbf{R}(x) \otimes \mathbf{S}(x),
\end{equation}
where $\mathbf{R}$ denotes surface reflectance and $\mathbf{S}$ represents illumination. Following recent works~\cite{rehit,cl3an,ambient6k}, we use reflectance $\mathbf{R}$ rather than classical Retinex albedo $\mathbf{A}$ to account for non-Lambertian effects (specular highlights, inter-reflections) common in real scenes. ReHiT~\cite{rehit} models this as perturbations $\hat{\mathbf{R}}$ and $\hat{\mathbf{L}}$ from ideal conditions, while RLN$^2$~\cite{cl3an} uses ambient-lit images as reflectance targets, both recognizing that $\mathbf{R}$ approximates perceptually stable appearance rather than strict Lambertian albedo.

Transforming to the logarithmic domain yields additive separability, $\log \mathbf{I}_{\text{norm}} = \log \mathbf{R} + \log \mathbf{S}$. Under the smoothness assumption, we estimate global lighting as the spatial average in log-space, $\log \hat{\mathbf{S}} = \mathbb{E}_{H,W}[\log(\mathbf{I}_{\text{norm}}+\varepsilon)]$, with pseudo-reflectance as the residual $\log \hat{\mathbf{R}} = \log(\mathbf{I}_{\text{norm}}+\varepsilon) - \log \hat{\mathbf{S}}$. Exponentiating yields $\hat{\mathbf{R}} = \exp(\log \hat{\mathbf{R}})$ and $\hat{\mathbf{S}} = \exp(\log \hat{\mathbf{S}})$, effectively isolating dominant lighting effects from material cues.

\textbf{Recombination and Normalization.}  
The pseudo-components are recombined and normalized:
\begin{equation}
\hat{\mathbf{I}} = \frac{\hat{\mathbf{R}} \otimes \hat{\mathbf{S}} - \min(\hat{\mathbf{R}} \otimes \hat{\mathbf{S}})}{\max(\hat{\mathbf{R}} \otimes \hat{\mathbf{S}}) - \min(\hat{\mathbf{R}} \otimes \hat{\mathbf{S}}) + \varepsilon},
\end{equation}
where $\min(\cdot)$ and $\max(\cdot)$ maintain valid radiometric relationships. As illustrated in Figure~\ref{fig:pan_pipeline}, this pipeline consistently improves shadow removal across diverse lighting conditions (Tables~\ref{tab:normalization_ablation_cross},~\ref{tab:normalization_ablation}), validating its effectiveness in handling multi-source illumination.
\begin{figure}[t]
\centering
\includegraphics[width=1\linewidth]{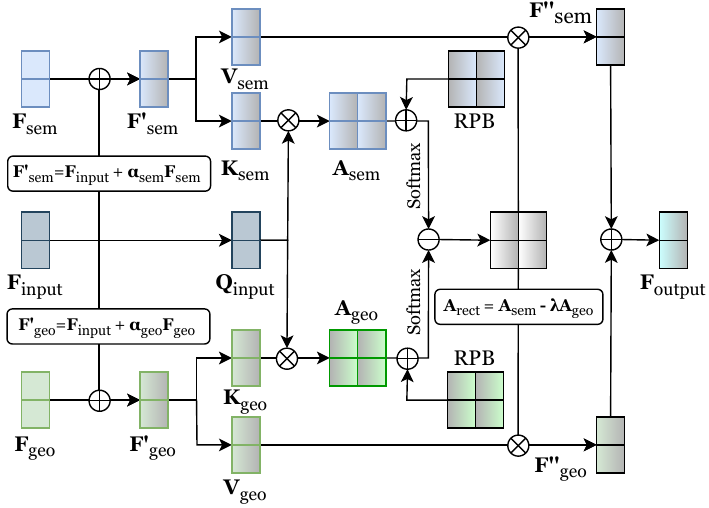}
\caption{\textbf{Overview of the proposed GSRA.} Semantic and geometric features are projected into modality-specific key–value spaces and queried with shared tokens. Rectification aligns the two modalities through soft attention balancing, resolving illumination inconsistencies while preserving geometric stability.}
\label{fig:GSRA_flow}
\vspace{-0.3cm}
\end{figure}

\subsection{Geometric Semantic Rectification Attention}
\label{sec:gsra}

While PAN provides illumination-consistent input representation, the network must reconcile complementary physical cues embedded in real-world images to accurately recover reflectance under varying lighting. Real-world scenes contain two distinct physically informed signals: \textit{geometric priors} (surface orientation, shading gradients, depth discontinuities) that encode local light-geometry interactions, and \textit{semantic embeddings} (object identity, material categories) that capture perceptually stable appearance across illumination changes. Under Lambertian reflection, local shading $I(x) = \rho(x) \cdot \mathbf{n}(x)^T \mathbf{d}$ depends on albedo $\rho$, normal $\mathbf{n}$, and light direction $\mathbf{d}$ \cite{horn1989shape}, making geometric features sensitive to lighting geometry. Conversely, semantic features remain stable—a red apple stays semantically "red apple" whether shadowed or sunlit—providing crucial context for resolving ambiguities in complex materials~\cite{rehit,cl3an}.

However, these modalities respond differently to illumination variations: geometric features are precise but noisy (sharp at shadow edges, smooth in uniformly lit regions), while semantic features are stable but spatially coarse. This \textit{modal alignment challenge} becomes critical for ambient lighting normalization task, where multiple overlapping light sources create complex interactions. Recent methods adopt diverse approaches: IFBlend~\cite{ambient6k} employs frequency-domain transformation for white lighting, RLN$^2$~\cite{cl3an} uses HSV-based hue mapping for RGB color lighting, while PromptNorm~\cite{promptnorm} integrates depth-derived geometric priors with prompt-guided normalization. Shadow removal methods like OmniSR~\cite{omnisr} and DenseSR~\cite{densesr} also leverage geometric-semantic priors, yet existing fusion strategies often fail to properly align complementary modal strengths—particularly when geometric precision and semantic stability must cooperate to disentangle overlapping illumination contributions.

We propose GSRA, which leverages Differential Attention~\cite{diff} to harmonize these physically grounded modalities. Differential Attention's subtraction structure naturally implements \textit{physically interpretable gating}: the operation $\mathbf{A}_{\text{sem}} - \lambda \cdot \mathbf{A}_{\text{geo}}$ rectifies semantic attention using geometric guidance, suppressing semantic over-smoothing at true illumination boundaries while preserving geometric precision. This produces modality-aware features that balance local geometric detail with global semantic stability, crucial for generalizing from single-light shadow removal to multi-source ALN scenarios.

\textbf{Multimodal Prior Injection.} 
Given shared query feature $\mathbf{F}_{\text{input}}$, geometry prior $\mathbf{F}_{\text{geo}}$ from DepthAnything-V2~\cite{depthganythingv2} (depth and normal maps), and semantic embedding $\mathbf{F}_{\text{sem}}$ from DINO-v2~\cite{dinov2}, we construct two complementary streams via prior injection:
{\small
\begin{equation}
\mathbf{F}_{\text{geo}}' = \mathbf{F}_{\text{input}} + \alpha_{\text{geo}}\mathbf{F}_{\text{geo}}, \quad
\mathbf{F}_{\text{sem}}' = \mathbf{F}_{\text{input}} + \alpha_{\text{sem}}\mathbf{F}_{\text{sem}},
\end{equation}
}

where learnable $\alpha_{\text{geo}}$ and $\alpha_{\text{sem}}$ control prior strength. This reinforces structure-preserving cues (shading continuity, edge orientation) in the geometric branch while adapting semantic context to illumination-dependent scene variations. We then generate modality-specific key-value pairs:
{\small
\begin{equation}
\{\mathbf{K}_{\text{geo}}, \mathbf{V}_{\text{geo}}\} = \mathcal{F}_{\text{geo}}(\mathbf{F}_{\text{geo}}'), \quad
\{\mathbf{K}_{\text{sem}}, \mathbf{V}_{\text{sem}}\} = \mathcal{F}_{\text{sem}}(\mathbf{F}_{\text{sem}}'),
\end{equation}
}
where $\mathcal{F}_{\text{geo}}(\cdot)$ and $\mathcal{F}_{\text{sem}}(\cdot)$ denote lightweight linear projections preserving modality characteristics.

\textbf{Differential Rectification.}
Following Differential Transformer (DT)~\cite{diff}, we apply differential rectification, but critically extend it to \textit{cross-modal attention}: while the original DT subtracts attention maps within a single self-attention mechanism to reduce noise, GSRA computes the difference \textit{between geometric and semantic modalities} to achieve physically grounded feature alignment. Using shared query $\mathbf{Q}_{\text{input}}$, we compute attention maps as $\mathbf{A}_{\text{geo}} = \text{Softmax}((\mathbf{Q}_{\text{input}}\mathbf{K}_{\text{geo}}^\top)/{\sqrt{d}} + \mathbf{B})$ and $\mathbf{A}_{\text{sem}} = \text{Softmax}((\mathbf{Q}_{\text{input}}\mathbf{K}_{\text{sem}}^\top)/{\sqrt{d}} + \mathbf{B})$, where $\mathbf{B}$ denotes relative position bias. Specifically, we rectify semantic attention using geometric guidance as $\mathbf{A}_{\text{rect}} = \mathbf{A}_{\text{sem}} - \lambda \cdot \mathbf{A}_{\text{geo}}$, where learnable $\lambda$ balances context-dependent illumination variation (high $\lambda$) versus geometric regularization (low $\lambda$). The fused output is obtained as $\mathbf{F}_{\text{output}} = \text{Concat}(\mathbf{A}_{\text{rect}}\mathbf{V}_{\text{geo}}, \mathbf{A}_{\text{rect}}\mathbf{V}_{\text{sem}})$, yielding features that harmonize local geometric precision with global semantic stability, achieving accurate shadow localization within the network, as shown in Figure~\ref{fig:featuremap}.

This formulation (illustrated in Figure~\ref{fig:GSRA_flow}) enables GSRA to maintain physically grounded geometric cues while selectively refining semantic responses. While recent shadow removal methods~\cite{shadowhack} also adopt differential attention mechanisms, GSRA distinguishes itself through \textit{explicit multi-modal prior injection}—directly incorporating geometric priors (depth, normals) and semantic embeddings (DINO-v2) to guide the rectification process. This design addresses the inherent challenge that traditional shadow removal benchmarks contain primarily single-light scenarios, whereas ambient lighting normalization demands robustness under multi-source indirect illumination and chromatic shifts. General restoration backbones~\cite{omnisr,densesr} similarly struggle with boundary preservation under such conditions.

\begin{table*}[t]
\small
\center
\scalebox{0.83}{
\begin{tabular}{l c c c c c c c c c c c}
\hline
\multirow{2}{*}{Method} & \multirow{2}{*}{Venue} & \multicolumn{2}{c}{ISTD Dataset} & \multicolumn{2}{c}{ISTD+ Dataset} & \multicolumn{2}{c}{INS Dataset} & \multicolumn{2}{c}{WSRD+ Dataset} & \multicolumn{2}{c}{Ambient6K Dataset}\\ \cmidrule(lr){3-4}\cmidrule(lr){5-6}\cmidrule(lr){7-8}\cmidrule(lr){9-10}\cmidrule(lr){11-12}
& & {PSNR$\uparrow$} & {SSIM$\uparrow$} & {PSNR$\uparrow$} & {SSIM$\uparrow$} & {PSNR$\uparrow$} & {SSIM$\uparrow$} & {PSNR$\uparrow$} & {SSIM$\uparrow$} & {PSNR$\uparrow$} & {SSIM$\uparrow$} \\  
\hline
DSC~\cite{DSC} & TPAMI 2019 & 29.00 & 0.944 & 25.66 & 0.956 & 29.05 & 0.940 & --- & --- & --- & --- \\ 
DHAN~\cite{cun2020towards} & AAAI 2020 & 29.11 & 0.954 & 25.66 & 0.956 & 27.84 & 0.963 & 22.39 & 0.796 & --- & ---\\ 

DC-ShadowNet~\cite{jin2021dc} & CVPR 2021 & 24.02 & 0.677 & 25.50 & 0.694 & --- & --- & 21.62 & 0.593 & 17.73 & 0.711\\
BMNet~\cite{zhu2022bijective} & CVPR 2022 & 28.53 & 0.952 & 32.22 & 0.965 & 27.90 & 0.958 & 24.75 & 0.816 & --- & ---\\ 
ShadowFormer~\cite{shadowformer} & AAAI 2023 & 29.90 & \cellcolor{gray!20}{0.960} & 31.39 & 0.946 & 28.62 & 0.963 & 25.44 & 0.820 & --- & ---\\ 
DMTN~\cite{liu2023decoupled} & TMM 2023 & 29.05 & {0.956} & 31.72 & 0.963 & 28.83 & 0.969 & --- & --- & --- & ---\\
ShadowDiffusion~\cite{guo2023shadowdiffusion} & CVPR 2023 & 30.09 & 0.918 & 31.08 & 0.950 & 29.12 & 0.966 & --- & --- & --- & --- \\ 
ShadowRefiner~\cite{shadowrefiner} & CVPRW 2024 & 28.75 & 0.916 & 31.03 & 0.928 & --- & --- & 26.04 & 0.827 & --- & --- \\ 
IFBlend~\cite{ambient6k} & ECCV 2024 & 28.55 & 0.906 & 30.87 &  0.916 & --- & --- & 25.79 & 0.809 & {21.44} & {0.819}\\ 
RLN$^2$-Lf \cite{cl3an} & ICCV 2025 & 28.77 & 0.914 & 31.02 & 0.930 & --- & --- & 25.84 & 0.821 & {21.71} & {0.825} \\
ReHiT \cite{rehit} & CVPRW 2025 & 28.81 & 0.914 & 31.16 & 0.925  & --- & --- & 26.15 & 0.826 & 19.98  & 0.798  \\
OmniSR \cite{omnisr} & AAAI 2025 & \cellcolor{gray!20}{30.45} & \cellcolor{gray!40}{0.964} & 33.34 & \cellcolor{gray!20}{0.970} & \cellcolor{gray!20}{30.38} & 0.973 & 26.07 & \cellcolor{gray!20}{0.835} & \cellcolor{gray!40}{23.01} & \cellcolor{gray!40}{0.830}  \\  
StableShadowDiffusion \cite{xu2024detailpreservinglatentdiffusionstable} & CVPR 2025 & --- & --- & \cellcolor{gray!70}{35.19} & \cellcolor{gray!20}{0.970} & \cellcolor{gray!40}{30.56} & \cellcolor{gray!40}{0.975} & \cellcolor{gray!20}{26.26} & 0.827 & --- & --- \\
DenseSR \cite{densesr} & ACMMM 2025 & \cellcolor{gray!40}{30.64} & \cellcolor{gray!70}{0.976} & \cellcolor{gray!20}{33.98} & \cellcolor{gray!70}{0.974} & \cellcolor{gray!70}{30.64} & \cellcolor{gray!70}{0.981} & \cellcolor{gray!40}{26.28} & \cellcolor{gray!40}{0.838} & \cellcolor{gray!20}{22.54} & \cellcolor{gray!20}{0.826} \\
\hline
\textbf{PhaSR (Ours)} & --- & \cellcolor{gray!70}{30.73} & \cellcolor{gray!20}{0.960} & \cellcolor{gray!40}{34.48} & 0.960 & \cellcolor{gray!20}{30.38} & 0.961 & \cellcolor{gray!70}{28.44} & \cellcolor{gray!70}{0.842} & \cellcolor{gray!70}{23.32} & \cellcolor{gray!70}{0.834} \\
\hline
\end{tabular}}
\caption{\textbf{Quantitative comparisons on shadow removal and ambient lighting normalization benchmarks.} 
We evaluate PhaSR on traditional SR datasets (ISTD~\cite{wang2018stacked}, ISTD+~\cite{le2019shadow}, INS~\cite{omnisr}, WSRD+~\cite{vasluianu2023wsrd}) and the challenging ambient lighting normalization benchmark (Ambient6K~\cite{ambient6k}), which contains multi-source white lighting and complex indirect illumination. 
PhaSR achieves SoTA results on WSRD+ and Ambient6K demonstrates robust generalization to ambient lighting scenarios.
Best results are highlighted as \colorbox{gray!70}{1st}, \colorbox{gray!40}{2nd}, and \colorbox{gray!20}{3rd}.}
\vspace{-0.2cm}
\label{tab:comparison_ISTD}
\end{table*}

\begin{figure}[t!]
\centering
\includegraphics[width=0.9\linewidth]{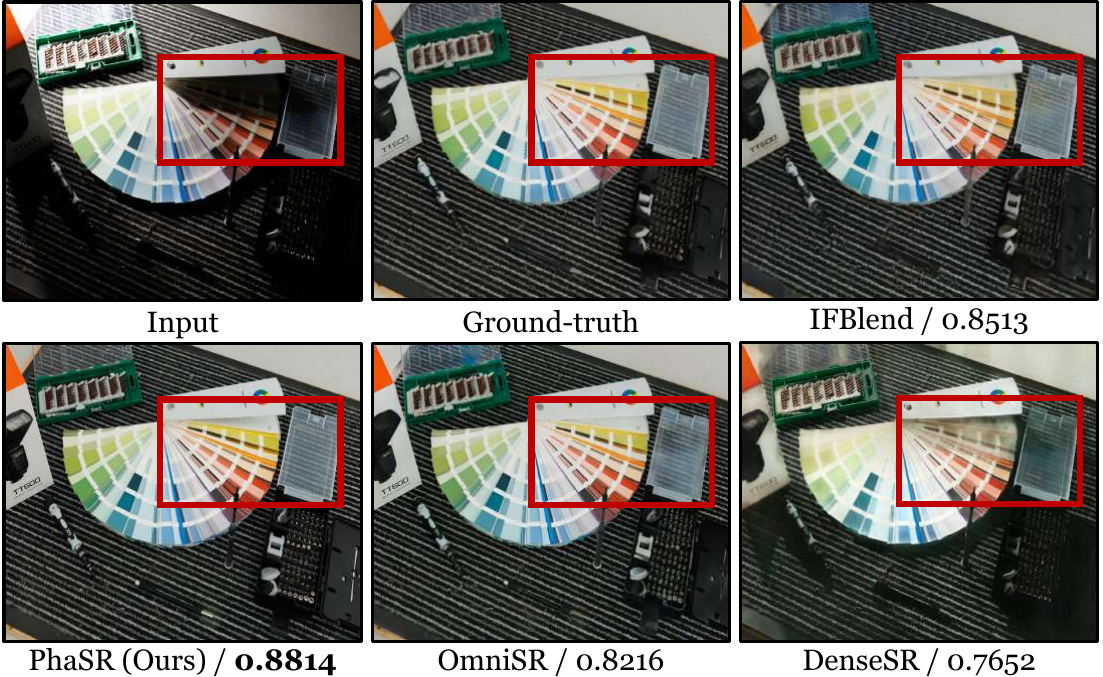}
\caption{\textbf{Qualitative comparison on Ambient6K~\cite{ambient6k}.} 
Ambient6K features multi-source white lighting and complex indirect illumination without shadow masks, requiring disentanglement of overlapping illumination contributions—substantially more challenging than single-light shadow removal. 
PhaSR effectively recovers ambient-normalized images while preserving material details. 
Best SSIM score is bolded in the figures.
}
\label{fig:Ambient6K} 
\vspace{-0.5cm}
\end{figure}

\begin{figure}[t!]
\centering
\includegraphics[width=1.0\linewidth]{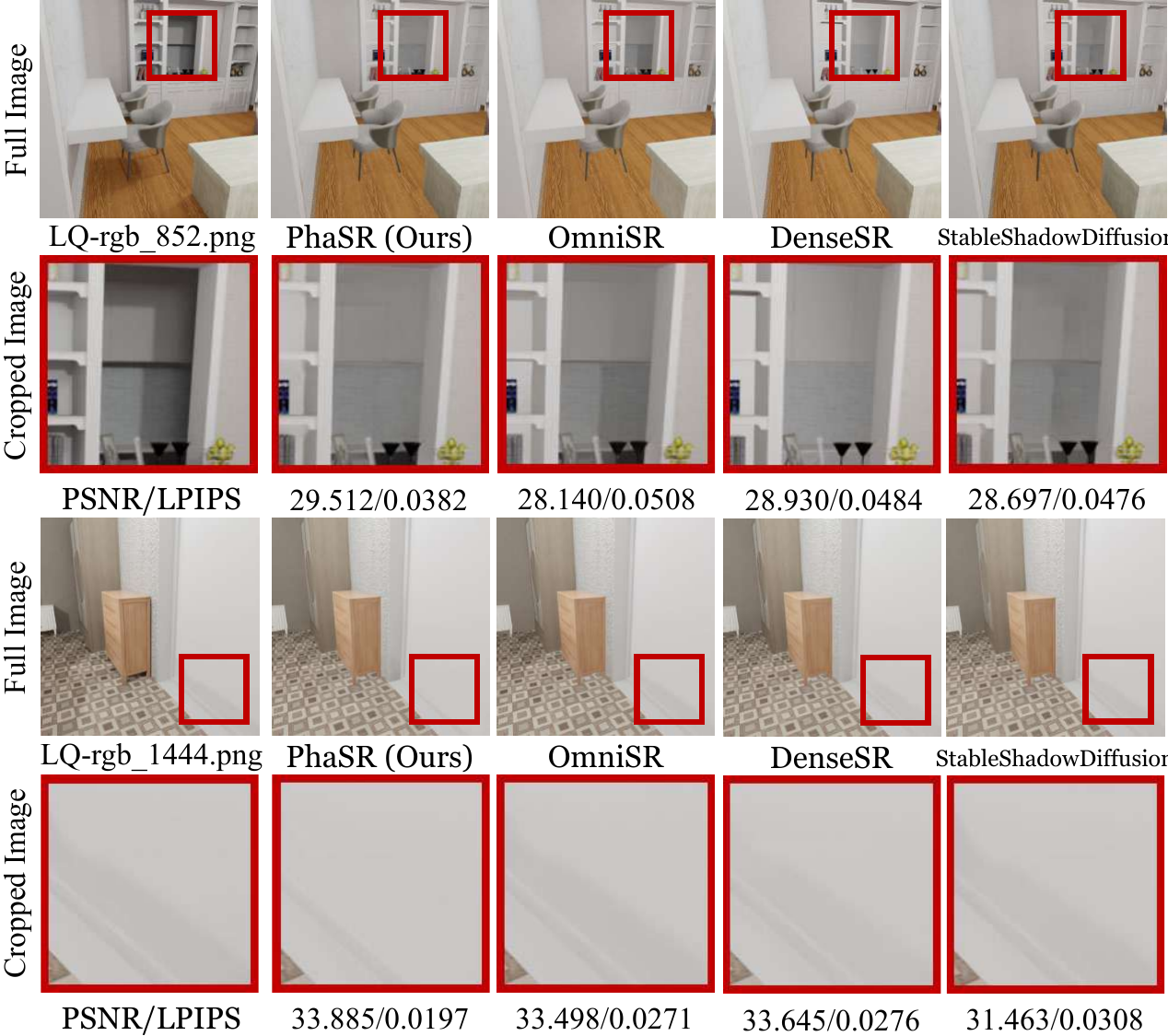}
\caption{\textbf{Results visualization on INS \cite{omnisr} dataset.} INS is a synthetic indoor dataset rendered with physically based lighting, featuring diverse materials and complex shadow interactions.}
\label{fig:INS} 
\vspace{-0.5cm}
\end{figure}

\subsection{Loss Functions}
\label{sec:loss_functions}
We supervise shadow-free image reconstruction using Charbonnier loss~\cite{zamir2020learning} for pixel-wise fidelity and SSIM loss for structural consistency. Given predicted output $\hat{\mathbf{I}}_{\text{pred}}$ and ground truth $\mathbf{I}_{\text{GT}}$, the losses are defined as:
{\footnotesize
\begin{equation}
\label{eq:losses}
\mathcal{L}_{\text{Charb}} = \sqrt{ \left\| \hat{\mathbf{I}}_{\text{pred}} - \mathbf{I}_{\text{GT}} \right\|_2^2 + \epsilon^2 }, \quad
\mathcal{L}_{\text{SSIM}} = 1 - \mathrm{SSIM}\!\left(\hat{\mathbf{I}}_{\text{pred}},\, \mathbf{I}_{\text{GT}}\right),
\end{equation}
}
where $\epsilon = 10^{-6}$ ensures numerical stability. The total objective is:$
\mathcal{L}_{\text{total}} = \lambda_{\text{Charb}}\,\mathcal{L}_{\text{Charb}} + \lambda_{\text{SSIM}}\,\mathcal{L}_{\text{SSIM}}$,
with $\lambda_{\text{Charb}} = 0.95$ and $\lambda_{\text{SSIM}} = 0.05$ balancing fidelity and perceptual quality.

\section{Experiment Results}
\label{sec:experimentresults}

\textbf{Datasets and Implementation Details.}
We evaluated our method on five datasets: ISTD \cite{wang2018stacked}, ISTD+ \cite{le2019shadow}, WSRD+ \cite{vasluianu2023wsrd}, INS \cite{omnisr} and Ambient6K \cite{ambient6k}. Following previous work \cite{fu2021auto,le2020shadow,shadowformer,omnisr}, we used $256 \times 256$ randomly cropped images and report PSNR and SSIM scores. For WSRD+, we used the evaluation data and code from the NTIRE 2024 Shadow Removal Challenge \cite{vasluianu2024ntire}.

Our model adopts a hierarchical architecture with base channel dimension $C=32$ and uniform depth configuration where each Transformer block ($\mathbf{N}_1$--$\mathbf{N}_7$) contains 2 layers. We used AdamW optimizer \cite{Kingma2015AdamAM} with $\beta_1$=0.9, $\beta_2$=0.999, $\epsilon=1 \times 10^{-8}$, batch size of 9, and 1400 epochs. The learning rate started at $2 \times 10^{-4}$ with cosine annealing. Standard augmentations, including random flipping and rotation were applied. All comparisons use the results reported in the original papers and hyperparameters.

\begin{figure*}[t!]
\centering
\includegraphics[width=1\linewidth]{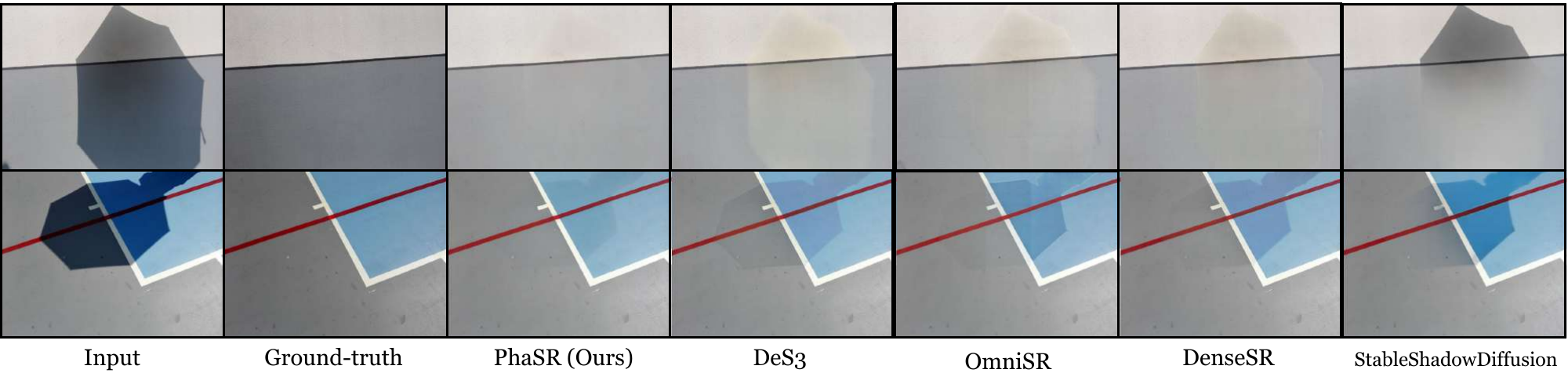}
\vspace{-0.4cm}
\caption{\textbf{Qualitative comparison with state-of-the-art methods on ISTD+~\cite{le2019shadow}.} 
ISTD+ is a color-corrected version of ISTD featuring outdoor scenes under natural single-light direct illumination. 
PhaSR effectively removes hard shadows while preserving fine-grained texture details and avoiding color distortion, demonstrating competitive performance against both mask-based and mask-free approaches.
}
\label{fig:comparison} 
\vspace{-0.4cm}
\end{figure*}


\subsection{Quantitative Results}
\label{sec:quantitative}

\begin{figure}[t!]
\centering
\includegraphics[width=1.0\linewidth]{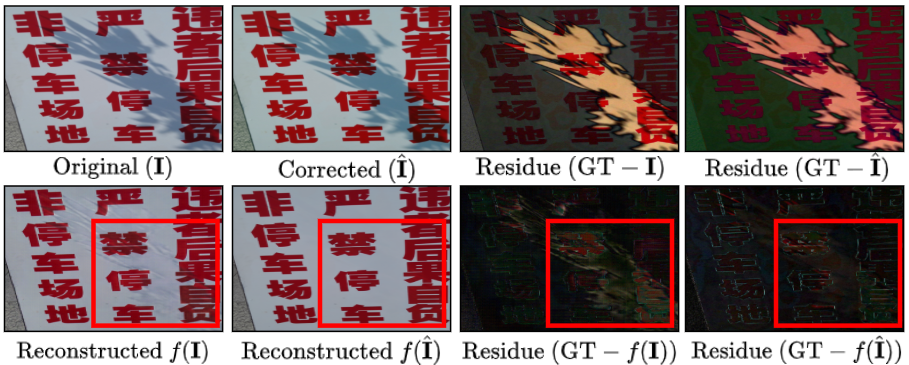}
\caption{\textbf{More examples of PAN.} For the real captured in ISTD+ \cite{le2019shadow} and WSTD+ \cite{vasluianu2023wsrd}, our method excels in removing complex indirect shadows and boundary sharpness. (The darker, the better for residue images, please zoom in for better view.)
}
\label{fig:PAN} 
\vspace{-0.3cm}
\end{figure}

We compare our method with SoTA image shadow removal methods, including DSC~\cite{DSC}, DC-ShadowNet~\cite{jin2021dc}, DHAN~\cite{cun2020towards}, BMNet~\cite{zhu2022bijective}, 
ShadowRefiner~\cite{shadowrefiner}, ShadowFormer~\cite{shadowformer}, DMTN~\cite{liu2023decoupled}, ShadowDiffusion~\cite{guo2023shadowdiffusion}, ReHiT~\cite{rehit}, OmniSR~\cite{omnisr}, StableShadowDiffusion~\cite{xu2024detailpreservinglatentdiffusionstable}, and DenseSR~\cite{densesr}, as shown in Tables \ref{tab:comparison_ISTD}. Note that methods requiring explicit shadow masks as input are excluded from our comparison, as all experiments are conducted under mask-free settings. This evaluation protocol better reflects real-world scenarios where automatic mask detection often fails due to varying lighting conditions, shadow softness, and complex scene compositions, making mask-free approaches more practical and robust for deployment. 

Qualitative results are presented in Figures \ref{fig:Ambient6K}, \ref{fig:INS}, and \ref{fig:comparison} for Ambient6K, INS, and ISTD+ datasets, respectively. PhaSR effectively preserves texture details while removing shadow artifacts across diverse scenarios. On the challenging Ambient6K dataset~\cite{ambient6k}, which involves complex multi-source illumination and diffuse indirect lighting beyond conventional shadow removal, PhaSR substantially outperforms dedicated ambient light normalization methods including IFBlend~\cite{ambient6k} and RLN$^2$-Lf~\cite{cl3an}. Interestingly, shadow removal methods that incorporate geometric or semantic priors, such as OmniSR~\cite{omnisr} and DenseSR~\cite{densesr}, also demonstrate improved performance on ambient light normalization, suggesting that structured prior knowledge benefits both shadow removal and complex lighting scenarios.

\subsection{Analysis of Physically Aligned Normalization}
\label{sec:pan_analysis}

We evaluate the proposed PAN to verify its impact on illumination consistency and restoration quality.  
As shown in Table~\ref{tab:normalization_ablation_cross}, PAN effectively reduces residual errors between shadowed and non-shadowed regions across diverse datasets.  
The improvement is most pronounced on outdoor scenes such as ISTD, achieving up to \textcolor{green!60!black}{26.4\%} error reduction, while even in diffuse indoor ambient lighting, which include multiple lighting sources and color shiftings (e.g., Ambient6K \cite{ambient6k}, CL3AN \cite{cl3an}), PAN maintains steady gains of 1–8\%.  
These results confirm PAN’s robustness in normalizing both direct and indirect illumination, yielding more uniform inputs for downstream processing.

When integrated as a plug-in module into various frameworks (OmniSR~\cite{omnisr}, DenseSR~\cite{densesr}, and PhaSR), PAN consistently improves PSNR/SSIM across all datasets (Table~\ref{tab:normalization_ablation}). Figure \ref{fig:PAN} shows that using PAN could minimise the residue of the corrected input. In summary, PAN serves as a parameter-free yet physically grounded normalization step that stabilizes input distributions and facilitates more reliable shadow removal under complex lighting.

\textbf{Comparison with traditional methods.}
As shown in Table~\ref{tab:PAN_comparison}, we compare PAN against classical parameter-free color correction methods including ACE~\cite{ace}, White-balance, White-Patch, and CIELab on WSRD+~\cite{vasluianu2023wsrd}.
Unlike these generic approaches that focus solely on global color consistency, PAN explicitly addresses shadow-specific challenges through log-domain decomposition and global-local illumination balancing.
This physically grounded design enables PAN to distinguish intrinsic reflectance from shadow-induced shading, achieving superior performance across all metrics—particularly in perceptual quality.


\subsection{Complexity Analysis}
As summarized in Table~\ref{tab:efficiency}, PhaSR achieves a favorable balance between accuracy and computational efficiency.
Despite integrating multimodal priors such as DepthAnything-V2 \cite{depthganythingv2} and DINO-V2 \cite{dinov2}, the proposed architecture maintains the lowest FLOPs (55.63 G) and the second smallest parameter count (18.95 M) among all compared models.
Thanks to its lightweight asymmetric decoder and modality-differential attention design, PhaSR runs in 87.9 ms per 640 × 480 image, faster than diffusion-based approaches like ShadowDiffusion \cite{shadowdiffusion} and StableShadowDiffusion \cite{xu2024detailpreservinglatentdiffusionstable}, while producing comparable or superior restoration quality.
These results demonstrate that the proposed physically aligned design not only enhances accuracy but also ensures high computational efficiency suitable for real-time or embedded deployment.

\begin{table*}[t]
\centering
\setlength\tabcolsep{5pt}
\small
\begin{minipage}[t]{0.58\textwidth}
\centering
\vspace{0pt}
\scalebox{0.784}{\begin{tabular}{lccccc}
\hline
& \multicolumn{4}{c}{\textbf{Residual Error (↓)}} \\
\hline
\textbf{Dataset}&\textbf{Subset} & \textbf{Original} & \textbf{Normalized} & \textbf{Improvement} & \textbf{Scene Type} \\
\hline
\multirow{2}{*}{ISTD \cite{wang2018stacked}} 
  & Train & 0.1525 & \textbf{0.1123} & \textcolor{green!60!black}{+26.4\%} & Real-world / Outdoor \\
  & Test  & 0.1199 & \textbf{0.0992} & \textcolor{green!60!black}{+17.3\%} & Real-world / Outdoor \\
\cmidrule(lr){1-6}
\multirow{2}{*}{INS \cite{omnisr}} 
  & Train & 0.0488 & \textbf{0.0471} & \textcolor{green!60!black}{+3.5\%} & Synthesized / Indoor \\
  & Test  & 0.0672 & \textbf{0.0643} & \textcolor{green!60!black}{+4.3\%} & Synthesized / Indoor \\
\cmidrule(lr){1-6}
\multirow{2}{*}{Ambient6K \cite{ambient6k}} 
  & Train & 0.1808 & \textbf{0.1793} & \textcolor{green!60!black}{+0.8\%} & Real-world / Indoor \\
  & Test  & 0.1870 & \textbf{0.1846} & \textcolor{green!60!black}{+1.3\%} & Real-world / Indoor \\
\cmidrule(lr){1-6}
\multirow{2}{*}{WSRD$+$ \cite{vasluianu2023wsrd}} 
  & Train & 0.1178 & \textbf{0.1134} & \textcolor{green!60!black}{+3.7\%} & Real-world / Indoor \\
  & Test  & 0.1223 & \textbf{0.1169} & \textcolor{green!60!black}{+4.4\%} & Real-world / Indoor \\
\cmidrule(lr){1-6}
\multirow{2}{*}{SRD \cite{qu2017deshadownet}} 
  & Train & 0.1382 & \textbf{0.1372} & \textcolor{green!60!black}{+0.7\%} & Real-world / Outdoor \\
  & Test  & 0.1689 & \textbf{0.1632} & \textcolor{green!60!black}{+3.4\%} & Real-world / Outdoor \\
\cmidrule(lr){1-6}
\multirow{4}{*}{CL3AN \cite{cl3an}} 
  & SH Train & 0.1566 & \textbf{0.1539} & \textcolor{green!60!black}{+1.7\%} & Real-world / Indoor \\
  & CR Train & 0.2899 & \textbf{0.2668} & \textcolor{green!60!black}{+8.0\%} & Real-world / Indoor \\
  & SH Test  & 0.3900 & \textbf{0.3762} & \textcolor{green!60!black}{+3.5\%} & Real-world / Indoor \\
  & CR Test  & 0.2930 & \textbf{0.2713} & \textcolor{green!60!black}{+7.4\%} & Real-world / Indoor \\
\hline
\end{tabular}

}

\raggedright
\hspace{-2mm}
\vspace{-1mm}

\caption{\textbf{Evaluation of PAN across diverse datasets.}
Residual error (↓) is the mean pixel-wise difference from shadow-free reference.
Leveraging proposed Normalization consistently improves results.}
\label{tab:normalization_ablation_cross}
\end{minipage}%
\hfill
\begin{minipage}[t]{0.39\textwidth}
\setlength\tabcolsep{3.5pt}
\small
\center
\scalebox{0.87}{
\begin{tabular}{lcc}
\toprule
\textbf{Dataset / Model} & \textbf{w/o PAN} & \textbf{w/ PAN} \\
\midrule
\multicolumn{3}{l}{\textbf{ISTD+ (Outdoor)}} \\
\midrule
OmniSR~\cite{omnisr}       & 30.45 / 0.964 & \textbf{30.67 / 0.969} \textcolor{green!60!black}{(+0.22/0.005)} \\
DenseSR~\cite{densesr}     & 30.64 / 0.976 & \textbf{30.69 / 0.979} \textcolor{green!60!black}{(+0.05/0.003)} \\
PhaSR (Ours)               & 30.58 / 0.954 & \textbf{30.73 / 0.960} \textcolor{green!60!black}{(+0.15/0.006)} \\
\midrule
\multicolumn{3}{l}{\textbf{WSRD+ (Indoor-Real, Single Lighting Sources)}} \\
\midrule
OmniSR~\cite{omnisr}       & 26.07 / 0.835 & \textbf{26.29 / 0.842} \textcolor{green!60!black}{(+0.22/0.007)} \\
DenseSR~\cite{densesr}     & 26.28 / 0.838 & \textbf{26.61 / 0.841} \textcolor{green!60!black}{(+0.33/0.003)} \\
PhaSR (Ours)               & 28.17 / 0.825 & \textbf{28.44 / 0.842} \textcolor{green!60!black}{(+0.27/0.017)} \\
\midrule
\multicolumn{3}{l}{\textbf{Ambient6K (Indoor-Real, Multiple Lighting Sources)}} \\
\midrule
OmniSR~\cite{omnisr}       & 23.01 / 0.830 & \textbf{23.25 / 0.832} \textcolor{green!60!black}{(+0.24/0.002)} \\
DenseSR~\cite{densesr}     & 22.54 / 0.826 & \textbf{22.78 / 0.830} \textcolor{green!60!black}{(+0.24/0.004)} \\
PhaSR (Ours)               & 22.98 / 0.821 & \textbf{23.32 / 0.834} \textcolor{green!60!black}{(+0.34/0.013)} \\
\bottomrule
\end{tabular}}
\vspace{-1mm}
\caption{\textbf{Ablation of PAN across SoTA prior-guided shadow removal models.}
Each cell reports PSNR/SSIM. 
Normalization improves both metrics.}
\label{tab:normalization_ablation}
\end{minipage}

\vspace{-0.35cm}
\end{table*}

\begin{table}
\setlength\tabcolsep{5pt}
\small
\center
\scalebox{0.8}{
\begin{tabular}{l c c c}
\hline
Model & Run-time & FLOPs & \#Params \\
\hline
DeS3 \cite{des3} & 254.6ms & 406.356 G & 67.444 M \\
ShadowFormer \cite{shadowformer} & \cellcolor{gray!70}43.7ms & \cellcolor{gray!40}64.602G & \cellcolor{gray!70}11.352M \\
DMTN \cite{liu2023decoupled} & \cellcolor{gray!40}82.6ms & 122.301G & \cellcolor{gray!20}22.830M \\
ShadowDiffusion \cite{shadowdiffusion} & 506.9ms & 174.658G & 55.376M \\
OmniSR \cite{omnisr} & 120.1ms & \cellcolor{gray!20}78.316G & 24.553M \\
StableShadowDiffusion \cite{xu2024detailpreservinglatentdiffusionstable}  & 452.8ms & 678.577G & 1329.824M \\
DenseSR \cite{densesr} & 124.6ms & 81.127G & 24.698M \\
\textbf{PhaSR (Ours)} & \cellcolor{gray!20}87.9ms & \cellcolor{gray!70}55.632G & \cellcolor{gray!40}18.949M \\
\hline
\end{tabular}}
\caption{\textbf{Comparison of efficiency and model complexity.} PhaSR achieves the second smallest parameter size and the lowest FLOPs among all compared models, while maintaining superior restoration accuracy, demonstrating its strong balance between effectiveness and efficiency.}
\label{tab:efficiency}
\vspace{-0.3cm}
\end{table}

\begin{table}
\setlength\tabcolsep{5pt}
\small
\center
\scalebox{0.8}{
\begin{tabular}{l c c c c c}
\hline
Metrics & ACE & White-balance & White-Patch & CIELab & \textbf{PAN (Ours)} \\
\hline
PSNR & 26.5843 & 27.1237 & 26.5123 & 25.4175 & 28.4421 \\
SSIM & 0.8106 & 0.8125 & 0.8063 & 0.8016 & 0.8418 \\
LPIPS & 0.6748& 0.0548 & 0.0654 & 0.0715 & 0.0469 \\
RMSE & 1.0840 & 0.9762 & 1.0743 & 1.1562 & 0.9452\\
\hline
\end{tabular}}
\caption{\textbf{Comparison of color correction methods on WSRD+ \cite{vasluianu2023wsrd}.} PAN achieves the best result among all compared methods.}
\label{tab:PAN_comparison}
\vspace{-0.7cm}
\end{table}

\subsection{Ablation Study}
To validate our design choices, we conduct ablation studies on PAN and GSRA across ISTD+ and WSRD+~\cite{vasluianu2024ntire} datasets (Table~\ref{tab:ablation}).

\textbf{Impact of PAN.} Removing PAN causes consistent performance drops, confirming that closed-form illumination normalization stabilizes input representations across lighting conditions. This validates PAN's role as a physically grounded preprocessing step that reduces chromatic bias before feature learning.

\textbf{Impact of GSRA.} Excluding GSRA yields larger declines, with the most significant drop on real-world data (WSRD+), demonstrating that cross-modal geometric-semantic alignment is critical under complex lighting where single-modality features are insufficient.

\textbf{Modality contributions.} Ablating geometric priors (depth, normals) reduces PSNR and SSIM, while removing semantic embeddings decreases performance. This asymmetry suggests semantic context provides stronger global consistency, while geometric cues offer complementary local precision—particularly evident in high-frequency shadow boundaries where geometry dominates.

\textbf{Rectification mechanism.} Disabling differential rectification ($\lambda=0$) degrades results, confirming that the subtraction operation $\mathbf{A}_{\text{rect}} = \mathbf{A}_{\text{sem}} - \lambda \cdot \mathbf{A}_{\text{geo}}$ effectively balances modal responses rather than naively fusing them.

These results validate that PAN and GSRA address complementary challenges: PAN achieves global illumination consistency, while GSRA resolves local modal conflicts, enabling physically coherent shadow removal under diverse lighting conditions. Notably, the larger performance gains on real-world indoor datasets compared to outdoor benchmarks confirm that explicit prior alignment becomes increasingly critical as the complexity of environmental lighting grows—from single-light direct shadows to multi-source ambient illumination.

\begin{table}[t]
\setlength\tabcolsep{4pt}
\small
\center
\scalebox{0.77}{
\begin{tabular}{c c c c}
\hline
\multirow{2}{*}{} & ISTD+ Dataset & WSRD+ Dataset\\
\cmidrule(lr){2-2}\cmidrule(lr){3-3}\cmidrule(lr){4-4}Configuration& {PSNR $\uparrow$} / {SSIM $\uparrow$} & {PSNR $\uparrow$} / {SSIM $\uparrow$}  \\
\hline
Full (PAN + GSRA) & \textbf{34.48 / 0.960} & \textbf{28.44 / 0.842}  \\
w/o PAN & 33.15 / 0.952 & 28.17 / 0.825 \\ 
w/o GSRA (Using cross-attention) & 32.56 / 0.934 & 26.92 / 0.820  \\
w/o Feature mixing & 33.24 / 0.954 & 27.68 / 0.836  \\
w/o Geometric prior & 33.52 / 0.956 & 27.85 / 0.838  \\
w/o Semantic prior & 33.38 / 0.955 & 27.71 / 0.837  \\
w/o Rectification ($\lambda=0$) & 32.89 / 0.951 & 27.32 / 0.832 \\
\hline
\end{tabular}}
\caption{\textbf{Ablation results of PhaSR components.} 
Each module contributes to illumination alignment and feature consistency. Removing GSRA leads to notable drops, validating the necessity of both physically aligned prior and cross-modal rectification.}
\label{tab:ablation}
\vspace{-0.5cm}
\end{table}

\section{Conclusion}
\label{sec:conclusion}

We present PhaSR, a framework for shadow removal through dual-level physically aligned prior integration. 
At the global level, Physically Aligned Normalization (PAN) performs closed-form illumination correction via log-domain Retinex decomposition, providing a stable foundation that enhances existing architectures across diverse lighting conditions. 
At the local level, Geometric-Semantic Rectification Attention (GSRA) extends differential attention to cross-modal alignment, harmonizing depth-derived geometry with semantic embeddings to resolve modal conflicts under varying illumination. 
Experiments demonstrate that PhaSR achieves competitive performance on standard shadow removal benchmarks while generalizing robustly to ambient lighting normalization scenarios. 

\clearpage
\setcounter{page}{1}
\maketitlesupplementary

\section*{Overview}
This supplementary material provides comprehensive details to support the main paper. The document is organized as follows:
\begin{itemize}
    \item \textbf{Section 1: Data Loading and Preprocessing} -- Details on depth-to-normal conversion, normal map normalization, and input preparation pipeline using DepthAnything-V2 and DINO-V2.
    
    \item \textbf{Section 2: Algorithm Description} -- Complete algorithmic specification of the PhaSR training pipeline, including physically aligned normalization (PAN), multi-scale feature extraction with prior integration, and geometric-semantic rectification attention (GSRA).
    
    \item \textbf{Section 3: Cross-Dataset Generalization} -- Evaluation of robustness across diverse lighting conditions through cross-dataset experiments (Ambient6K $\leftrightarrow$ ISTD), demonstrating PhaSR's superior generalization from single-source outdoor shadows to multi-source indoor ambient lighting.
    
    \item \textbf{Section 4: Additional Visual Comparisons} -- Extensive qualitative results on ISTD+, WSRD+, INS, and Ambient6K datasets, demonstrating PhaSR's effectiveness across diverse shadow removal scenarios.
    
    \item \textbf{Section 5: Additional Feature Map Comparison} -- Intermediate feature map visualization comparing PhaSR with OmniSR and DenseSR, validating the effectiveness of physically aligned prior propagation.
    
    \item \textbf{Section 6: Failure Case Study} -- Analysis of challenging scenarios including dark intrinsic materials and specular surfaces, discussing limitations and future directions.
    
    \item \textbf{Section 7: Network Architecture Details} -- Complete architecture specification with layer-by-layer breakdown of input/output dimensions and operations.
\end{itemize}

\vspace{5mm}

\section{Data Loading and Preprocessing}

PhaSR requires four inputs: (1) RGB image, (2) depth map, (3) normal map, and (4) semantic feature map. Depth and semantic features are extracted using pretrained DepthAnything-v2~\cite{depthganythingv2} and DINO-v2~\cite{dinov2} models, following common practice in recent shadow removal literature~\cite{omnisr,densesr}. Normal maps are derived from depth using standard geometric conversion.

\textbf{Depth-to-Normal Conversion.} 
Given depth map $\mathbf{D} \in \mathbb{R}^{H \times W}$ and camera field-of-view (FOV = 60°), we first compute camera intrinsics:
\begin{equation}
    f = \frac{W}{2 \tan(\text{FOV}_{\text{rad}} / 2)}, \quad
    c_x = \frac{W - 1}{2}, \quad
    c_y = \frac{H - 1}{2},
\end{equation}
where $\text{FOV}_{\text{rad}} = \text{FOV}_{\text{deg}} \times \pi / 180$. Each pixel $(x, y)$ with depth $z = \mathbf{D}[y, x]$ is unprojected to 3D coordinates via the pinhole camera model:
\begin{equation}
    x_{\text{3d}} = \frac{(x - c_x) z}{f}, \quad
    y_{\text{3d}} = \frac{(y - c_y) z}{f}.
\end{equation}
The resulting 3D point cloud is then converted to surface normals via spatial gradients, yielding $\mathbf{N} \in \mathbb{R}^{H \times W \times 3}$.

\textbf{Normal Map Normalization.} 
Raw normal maps $\mathbf{n}_{\text{raw}} \in [0, 1]^3$ from depth estimation are rescaled to $[-1, 1]$ and $\ell_2$-normalized:
\begin{equation}
    \mathbf{n}_{\text{rescaled}} = 2\mathbf{n}_{\text{raw}} - 1, \quad
    \mathbf{n}_{\text{normalized}} = \frac{\mathbf{n}_{\text{rescaled}}}{\|\mathbf{n}_{\text{rescaled}}\|_2 + \epsilon},
\end{equation}
where $\epsilon = 10^{-20}$ ensures numerical stability. This produces unit-length normal vectors suitable for geometric feature extraction in GSRA.

\section{Algorithm Description}
\label{sec:algorithm}

We provide a detailed algorithmic description of PhaSR in Algorithm~\ref{alg:phasr}, which illustrates the complete training pipeline including physically aligned normalization, multi-scale feature extraction with prior integration, and geometric-semantic rectification attention.

\begin{algorithm}[t]
\caption{PhaSR Training Algorithm}
\label{alg:phasr}
\scalebox{0.82}{
\begin{minipage}{1.22\linewidth}
\begin{algorithmic}[1]
\REQUIRE Shadow image $\mathbf{I} \in \mathbb{R}^{H \times W \times 3}$, ground truth $\mathbf{I}_{\mathrm{GT}}$
\ENSURE Predicted shadow-free image $\hat{\mathbf{I}}$

\STATE \textbf{Stage 1: Physically Aligned Normalization (PAN)}
\STATE Gray-world: $\mathbf{I}_{\mathrm{norm}} = \mathbf{I} \cdot \frac{\mathbb{E}[\mathbf{I}]}{\mathbb{E}_c[\mathbf{I}] + \epsilon}$
\STATE Log-domain: $\log \hat{\mathbf{S}} = \mathbb{E}_{H,W}[\log(\mathbf{I}_{\mathrm{norm}} + \epsilon)]$, $\log \hat{\mathbf{R}} = \log(\mathbf{I}_{\mathrm{norm}} + \epsilon) - \log \hat{\mathbf{S}}$
\STATE Recombine: $\hat{\mathbf{I}} = \frac{\hat{\mathbf{R}} \otimes \hat{\mathbf{S}} - \min(\hat{\mathbf{R}} \otimes \hat{\mathbf{S}})}{\max(\hat{\mathbf{R}} \otimes \hat{\mathbf{S}}) - \min(\hat{\mathbf{R}} \otimes \hat{\mathbf{S}}) + \epsilon}$
\STATE \hspace{3.5em} where $\hat{\mathbf{R}} = \exp(\log \hat{\mathbf{R}})$, $\hat{\mathbf{S}} = \exp(\log \hat{\mathbf{S}})$

\STATE \textbf{Stage 2: Prior Extraction}
\STATE Extract features: $\mathbf{F}_{\mathrm{D}}^{(i)} = \mathrm{DINOV2}(\mathbf{I})$ for $i = 0, 1, 2, 3$
\STATE Extract depth and normals: $\mathbf{D} = \mathrm{DepthV2}(\mathbf{I})$, $\mathbf{N} = \nabla \mathbf{D}$

\STATE \textbf{Stage 3: Encoder with Prior Fusion}
\STATE Input projection: $\mathbf{y}_0 = \mathrm{InputProj}([\hat{\mathbf{I}}, \mathbf{D}_z])$
\FOR{$\ell = 0, \ldots, 3$}
    \STATE Project DINO: $\mathbf{F}_{\mathrm{d}}^{(\ell)} = \mathrm{Proj}(\mathrm{Up}(\mathbf{F}_{\mathrm{D}}^{(\ell)}))$
    \STATE Fuse: $\mathbf{y}_\ell = \mathbf{y}_\ell + \alpha_\ell \mathbf{F}_{\mathrm{d}}^{(\ell)}$
    \IF{$\ell < 3$}
        \STATE Encode: $\mathbf{c}_\ell = \mathrm{TEB}_\ell(\mathbf{y}_\ell, \mathbf{F}_{\mathrm{D}}^{(\ell)}, \mathbf{D}^{(\ell)}, \mathbf{N}^{(\ell)})$
        \STATE Downsample: $\mathbf{y}_{\ell+1} = \mathrm{Down}(\mathbf{c}_\ell)$
    \ENDIF
\ENDFOR

\STATE \textbf{Stage 4: Bottleneck}
\STATE Concatenate scales: $\mathbf{F}_{\mathrm{cat}} = \mathrm{Conv}([\mathbf{F}_{\mathrm{D}}^{(0)}, \mathbf{F}_{\mathrm{D}}^{(1)}, \mathbf{F}_{\mathrm{D}}^{(2)}, \mathbf{F}_{\mathrm{D}}^{(3)}])$
\STATE Bottleneck: $\mathbf{c}_3 = \mathrm{PATB}([\mathbf{y}_3 + \alpha_3 \mathbf{F}_{\mathrm{d}}^{(3)}, \mathbf{F}_{\mathrm{cat}}], \mathbf{F}_{\mathrm{D}}^{(3)}, \mathbf{D}^{(3)}, \mathbf{N}^{(3)})$

\STATE \textbf{Stage 5: Decoder with GSRA}
\FOR{$\ell = 2, 1, 0$}
    \STATE Upsample and skip: $\mathbf{u}_\ell = [\mathrm{Up}(\mathbf{c}_{\ell+1}), \mathbf{c}_\ell]$
    \STATE Feature mixing: $\mathbf{F}'_{\mathrm{g}} = \mathbf{u}_\ell + \alpha_{\mathrm{g}} \mathbf{F}_{\mathrm{g}}^{(\ell)}$, $\mathbf{F}'_{\mathrm{s}} = \mathbf{u}_\ell + \alpha_{\mathrm{s}} \mathbf{F}_{\mathrm{s}}^{(\ell)}$
    \STATE Generate KV: $\mathbf{K}_{\mathrm{g}}, \mathbf{V}_{\mathrm{g}} = \mathcal{F}_{\mathrm{g}}(\mathbf{F}'_{\mathrm{g}})$; $\mathbf{K}_{\mathrm{s}}, \mathbf{V}_{\mathrm{s}} = \mathcal{F}_{\mathrm{s}}(\mathbf{F}'_{\mathrm{s}})$
    \STATE Compute attention: $\mathbf{A}_{\mathrm{g}} = \mathrm{SoftMax}(\mathbf{Q} \mathbf{K}_{\mathrm{g}}^{\top} / \sqrt{d} + \mathbf{B})$
    \STATE \hspace{3.5em} $\mathbf{A}_{\mathrm{s}} = \mathrm{SoftMax}(\mathbf{Q} \mathbf{K}_{\mathrm{s}}^{\top} / \sqrt{d} + \mathbf{B})$
    \STATE Rectify: $\mathbf{A}_{\mathrm{r}} = \mathbf{A}_{\mathrm{s}} - \lambda^{(\ell)} \mathbf{A}_{\mathrm{g}}$
    \STATE Aggregate: $\mathbf{F}_{\mathrm{o}} = [\mathbf{A}_{\mathrm{r}} \mathbf{V}_{\mathrm{g}}, \mathbf{A}_{\mathrm{r}} \mathbf{V}_{\mathrm{s}}]$
    \STATE Decode: $\mathbf{c}_\ell = \mathrm{TDB}_\ell(\mathbf{F}_{\mathrm{o}}, \mathbf{F}_{\mathrm{D}}^{(\ell)}, \mathbf{D}^{(\ell)}, \mathbf{N}^{(\ell)})$
\ENDFOR

\STATE \textbf{Stage 6: Output and Loss}
\STATE Output: $\hat{\mathbf{I}} = \mathrm{OutProj}(\mathbf{c}_0) + \mathbf{I}$
\STATE Loss: $\mathcal{L} = \lambda_{\mathrm{C}} \sqrt{\|\hat{\mathbf{I}} - \mathbf{I}_{\mathrm{GT}}\|_2^2 + \epsilon^2} + \lambda_{\mathrm{S}} (1 - \mathrm{SSIM}(\hat{\mathbf{I}}, \mathbf{I}_{\mathrm{GT}}))$
\end{algorithmic}
\end{minipage}
}
\end{algorithm}

\section{Cross-Dataset Generalization}

To evaluate robustness across diverse lighting conditions, we conduct cross-dataset experiments where models trained on one dataset are directly tested on another without fine-tuning. As shown in Table~\ref{tab:cross_dataset}, PhaSR demonstrates competitive generalization capability in both directions.

\textbf{Ambient6K $\rightarrow$ ISTD.} When trained on complex multi-source indoor lighting and tested on single-light outdoor shadows, PhaSR consistently outperforms both OmniSR \cite{omnisr} and ShadowFormer \cite{shadowformer}, achieving improvements of +1.46 dB and +3.32 dB in PSNR respectively. These results suggest that our physically aligned design—global illumination normalization via PAN and local geometric-semantic rectification via GSRA—may contribute to effective generalization from complex to simpler lighting scenarios.

\textbf{ISTD $\rightarrow$ Ambient6K.} The reverse direction poses greater challenges, as models trained on direct single-light shadows must adapt to multi-source ambient illumination with overlapping light contributions and chromatic shifts. PhaSR maintains strong performance, outperforming competing methods by +2.33 dB over OmniSR and +4.90 dB over ShadowFormer. Notably, while all methods experience performance drops compared to in-domain training, PhaSR exhibits relatively smaller degradation, suggesting that explicit physical alignment may be associated with more robust feature learning across illumination distributions.

These results indicate that PhaSR's dual-level alignment strategy—closed-form illumination correction followed by cross-modal prior rectification—provides a design that generalizes effectively across datasets, from single-source outdoor shadows to multi-source indoor ambient lighting.

\begin{table}[h]
\centering
\caption{\textbf{Cross-dataset generalization evaluation.} Models trained on one dataset and tested on another to evaluate robustness across different lighting conditions.}
\label{tab:cross_dataset}
\resizebox{\columnwidth}{!}{
\begin{tabular}{l|cc|cc}
\toprule
\multirow{2}{*}{Method} & \multicolumn{2}{c|}{Ambient6K $\rightarrow$ ISTD} & \multicolumn{2}{c}{ISTD $\rightarrow$ Ambient6K} \\
& PSNR$\uparrow$ & SSIM$\uparrow$ & PSNR$\uparrow$ & SSIM$\uparrow$ \\
\midrule
ShadowFormer~\cite{shadowformer} & 24.32 & 0.872 & 16.25 & 0.671 \\
OmniSR~\cite{omnisr} & 26.18 & 0.901 & 18.82 & 0.733 \\
\rowcolor{gray!20}
\textbf{PhaSR (Ours)} & \textbf{27.64} & \textbf{0.923} & \textbf{21.15} & \textbf{0.798} \\
\midrule
\multicolumn{5}{l}{\textit{Reference: In-domain performance}} \\
ShadowFormer (ISTD) & 29.90 & 0.960 & — & — \\
OmniSR (ISTD) & 30.45 & 0.964 & — & — \\
\rowcolor{gray!20}
\textbf{PhaSR} (ISTD) & \textbf{30.73} & \textbf{0.960} & — & — \\
\midrule
ShadowFormer (Ambient6K) & — & — & 19.02 & 0.750 \\
OmniSR (Ambient6K) & — & — & 23.01 & 0.830 \\
\rowcolor{gray!20}
\textbf{PhaSR} (Ambient6K) & — & — & \textbf{23.32} & \textbf{0.834} \\
\bottomrule
\end{tabular}
}
\end{table}

\begin{figure}[t!]
\centering
\includegraphics[width=1.0\linewidth]{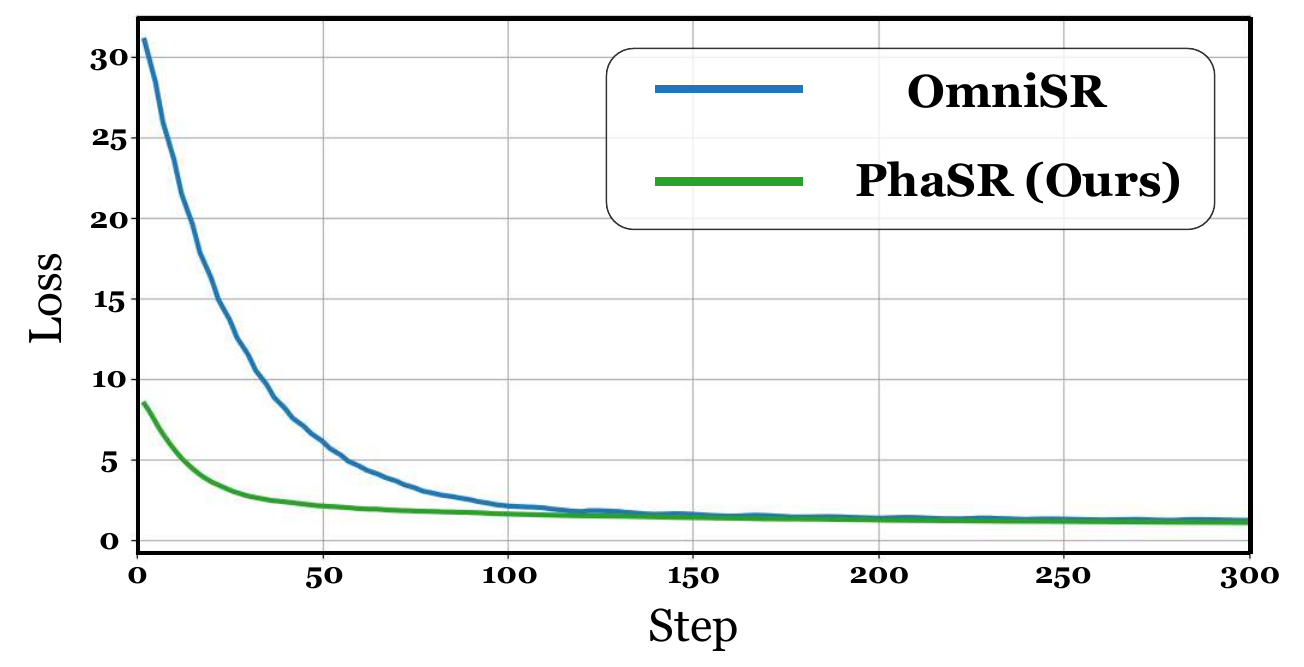}
\caption{\textbf{Training error of OmniSR \cite{omnisr} with and our method on WSRD+ dataset \cite{vasluianu2023wsrd}.} PhaSR yields an accelerated rate of error reduction.}
\label{fig:optimazationcurve} 
\vspace{-5mm}
\end{figure}

\section{Additional Visual Comparisons}
\label{sec:additional_visual}

\begin{figure*}[t!]
\centering
\includegraphics[width=1.0\linewidth]{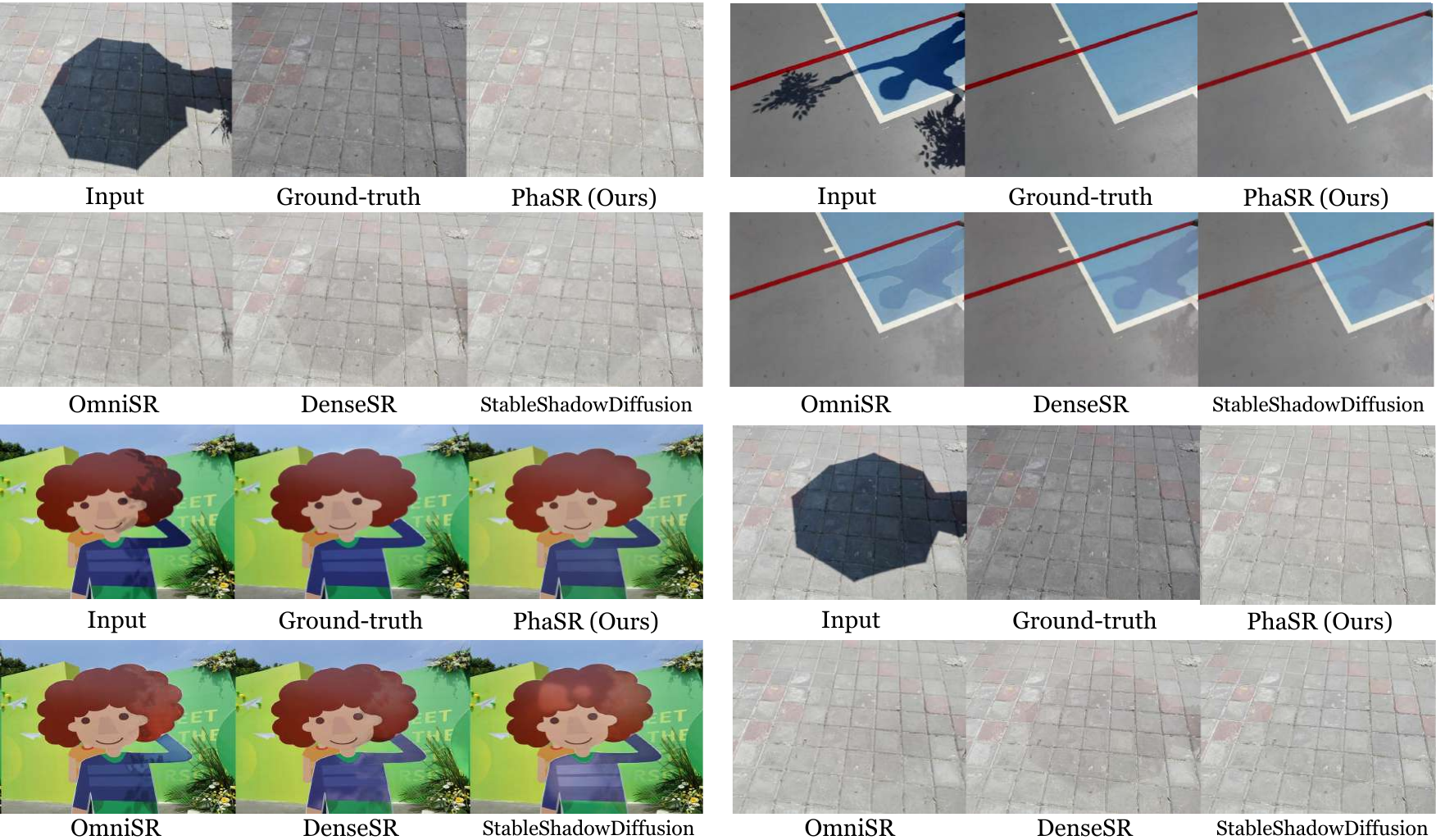}
\caption{\textbf{Additional visual comparisons on ISTD+~\cite{le2019shadow}.} PhaSR achieves superior shadow removal with sharper boundaries and better texture preservation compared to state-of-the-art methods.}
\label{fig:ISTD_supplementary} 
\vspace{-3mm}
\end{figure*}

\begin{figure*}[t!]
\centering
\includegraphics[width=1.0\linewidth]{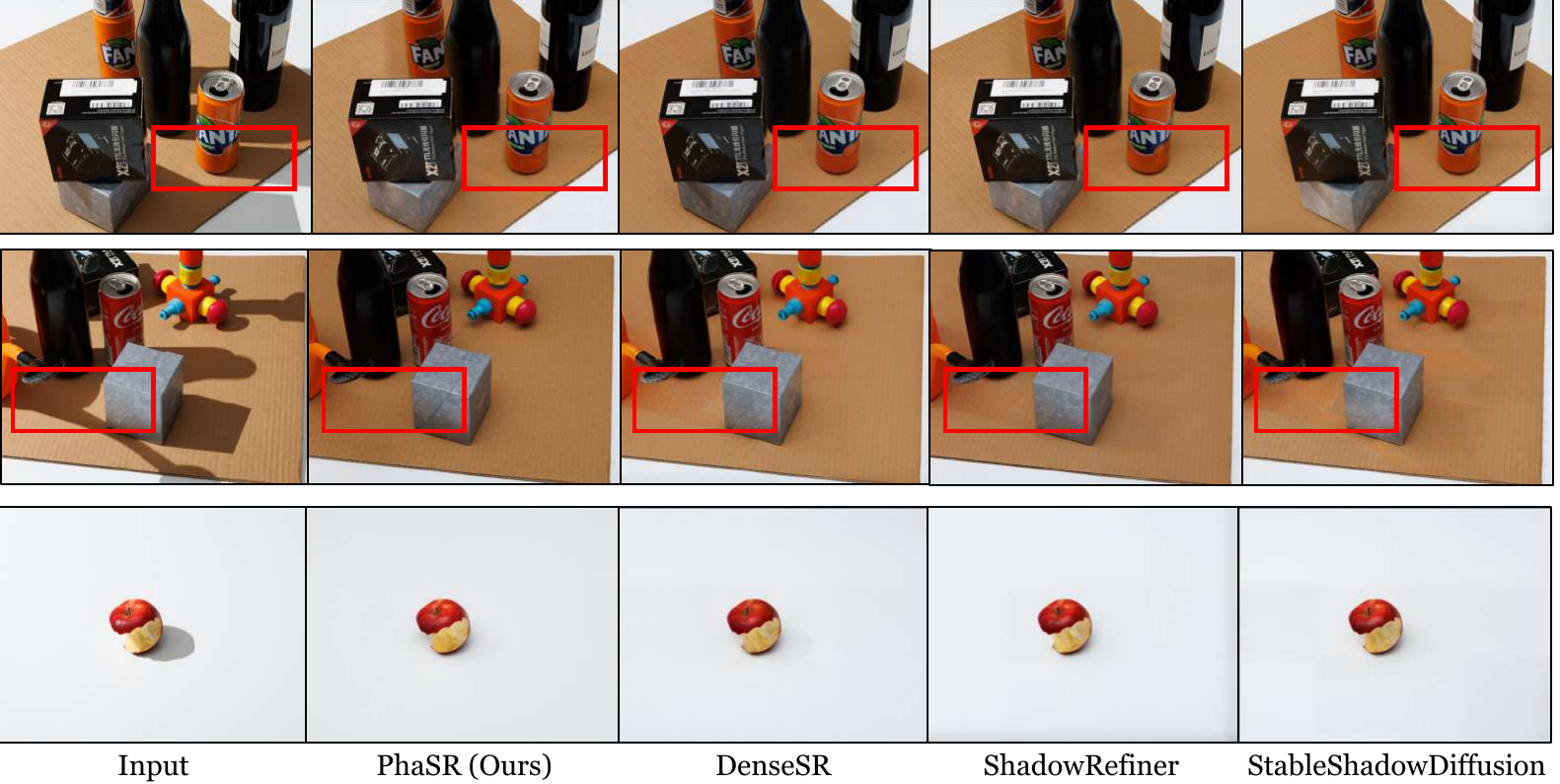}
\caption{\textbf{Additional visual comparisons on WSRD+~\cite{vasluianu2023wsrd}.} Our method effectively handles high-resolution indoor scenes with complex single-source lighting while maintaining photorealistic quality.}
\label{fig:WSRD_supplementary} 
\vspace{-3mm}
\end{figure*}

\begin{figure*}[t!]
\centering
\includegraphics[width=1.0\linewidth]{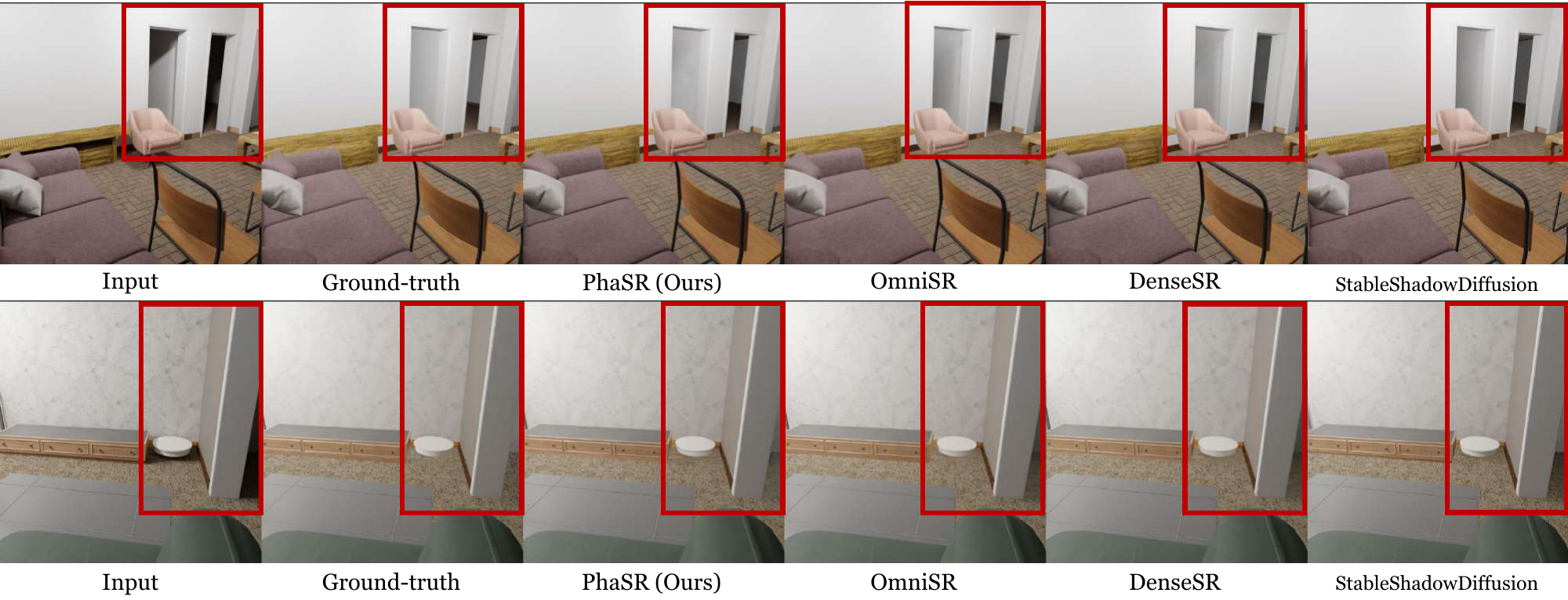}
\caption{\textbf{Additional visual comparisons on INS~\cite{omnisr}.} PhaSR demonstrates robust generalization to synthesized indoor scenes with indirect illumination and soft shadows.}
\label{fig:INS_supplementary} 
\vspace{-3mm}
\end{figure*}

\begin{figure*}[t!]
\centering
\includegraphics[width=1.0\linewidth]{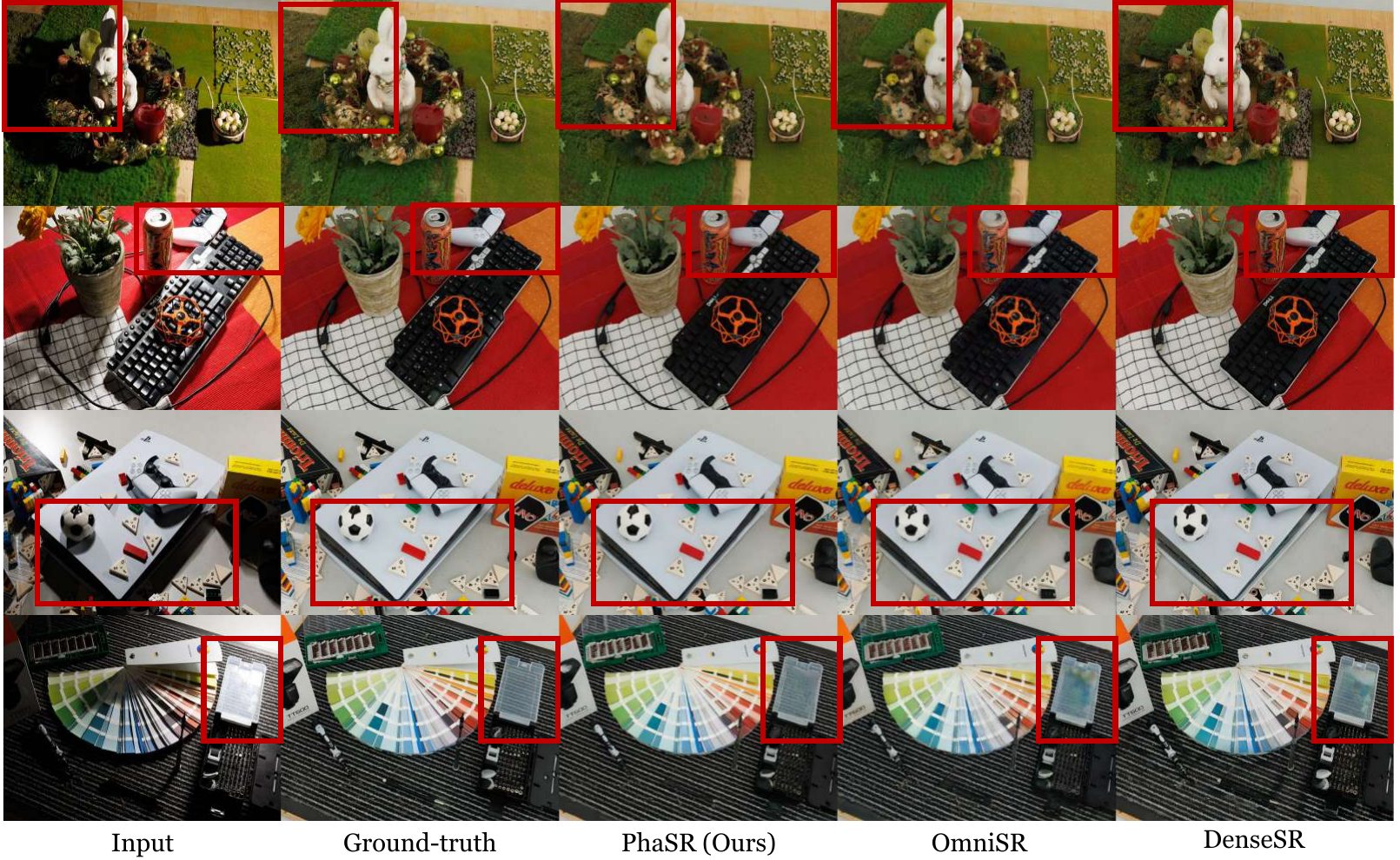}
\caption{\textbf{Additional visual comparisons on Ambient6K~\cite{ambient6k}.} PhaSR shows superior generalization to complex multi-source illumination and diffuse indirect lighting beyond conventional shadow removal, outperforming both ambient light normalization and shadow removal methods.}
\label{fig:A6K_supplementary} 
\vspace{-3mm}
\end{figure*}

We provide additional qualitative results to demonstrate the effectiveness of PhaSR across diverse shadow removal scenarios. Figures~\ref{fig:ISTD_supplementary}, \ref{fig:WSRD_supplementary}, \ref{fig:INS_supplementary}, and \ref{fig:A6K_supplementary} show comprehensive comparisons with state-of-the-art methods on ISTD+~\cite{le2019shadow}, WSRD+~\cite{vasluianu2023wsrd}, INS~\cite{omnisr}, and Ambient6K~\cite{ambient6k} datasets, respectively.

As shown in Figure~\ref{fig:ISTD_supplementary}, PhaSR generally recovers sharper shadow boundaries and preserves texture details compared to competing methods on real-world outdoor scenes. The proposed PAN effectively normalizes illumination variations, while GSRA resolves geometric-semantic ambiguities, leading to cleaner shadow-free results.

Figure~\ref{fig:WSRD_supplementary} demonstrates PhaSR's strong performance on high-resolution indoor scenes with complex single-source lighting. Compared to OmniSR~\cite{omnisr} and DenseSR~\cite{densesr}, which show some smoothing or color artifacts in certain regions, our method maintains photorealistic appearance while effectively removing shadow artifacts.

In Figure~\ref{fig:INS_supplementary}, we observe that PhaSR performs well on challenging synthesized indoor scenarios with indirect lighting and soft shadows. The physically aligned normalization appears to facilitate robust generalization across diverse illumination conditions, while the cross-modal attention mechanism effectively disentangles reflectance from shading.

Figure~\ref{fig:A6K_supplementary} further validates PhaSR's generalization capability on the challenging Ambient6K dataset, which features complex multi-source illumination and diffuse indirect lighting that goes beyond conventional shadow removal. Our method outperforms both dedicated ambient light normalization methods (IFBlend~\cite{ambient6k}) and shadow removal methods (OmniSR~\cite{omnisr}, DenseSR~\cite{densesr}). These results are consistent with the hypothesis that physically aligned design may facilitate handling diverse real-world lighting conditions.

\begin{figure}[h]
    \centering
    \setlength{\tabcolsep}{1pt} 
    \renewcommand{\arraystretch}{0.5}
    \vspace{-3mm}
    \begin{tabular}{ccccc}
        \includegraphics[width=0.19\linewidth]{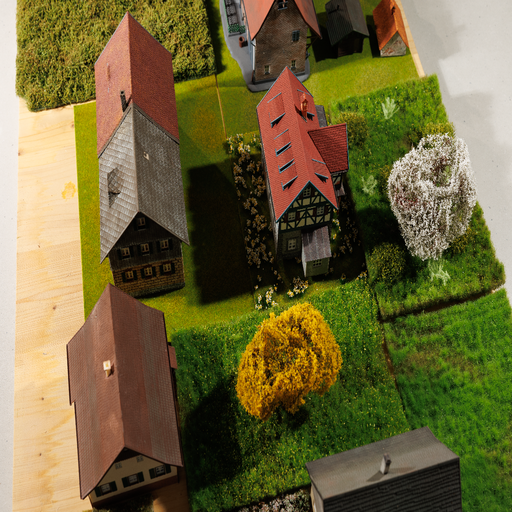} &
        \includegraphics[width=0.19\linewidth]{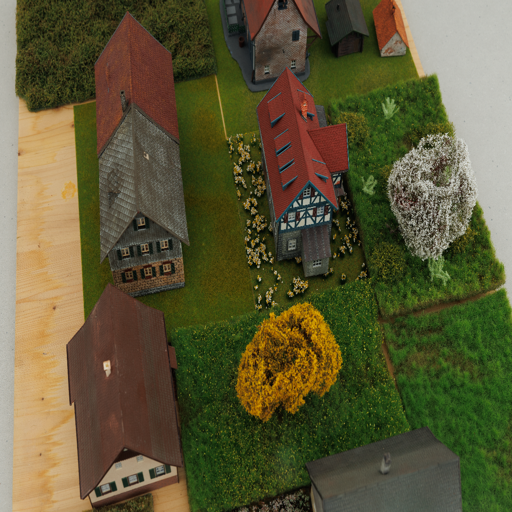} &
        \includegraphics[width=0.19\linewidth]{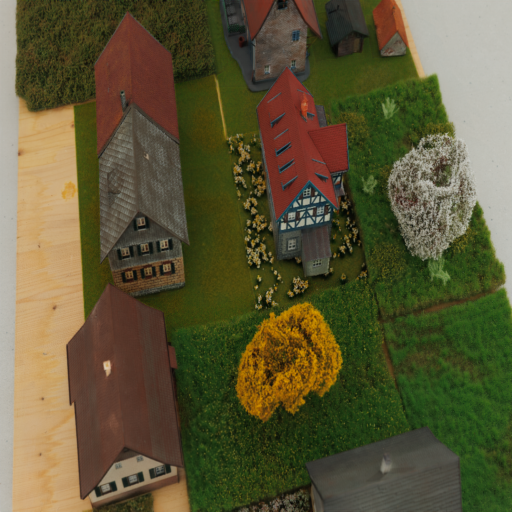} &
        \includegraphics[width=0.19\linewidth]{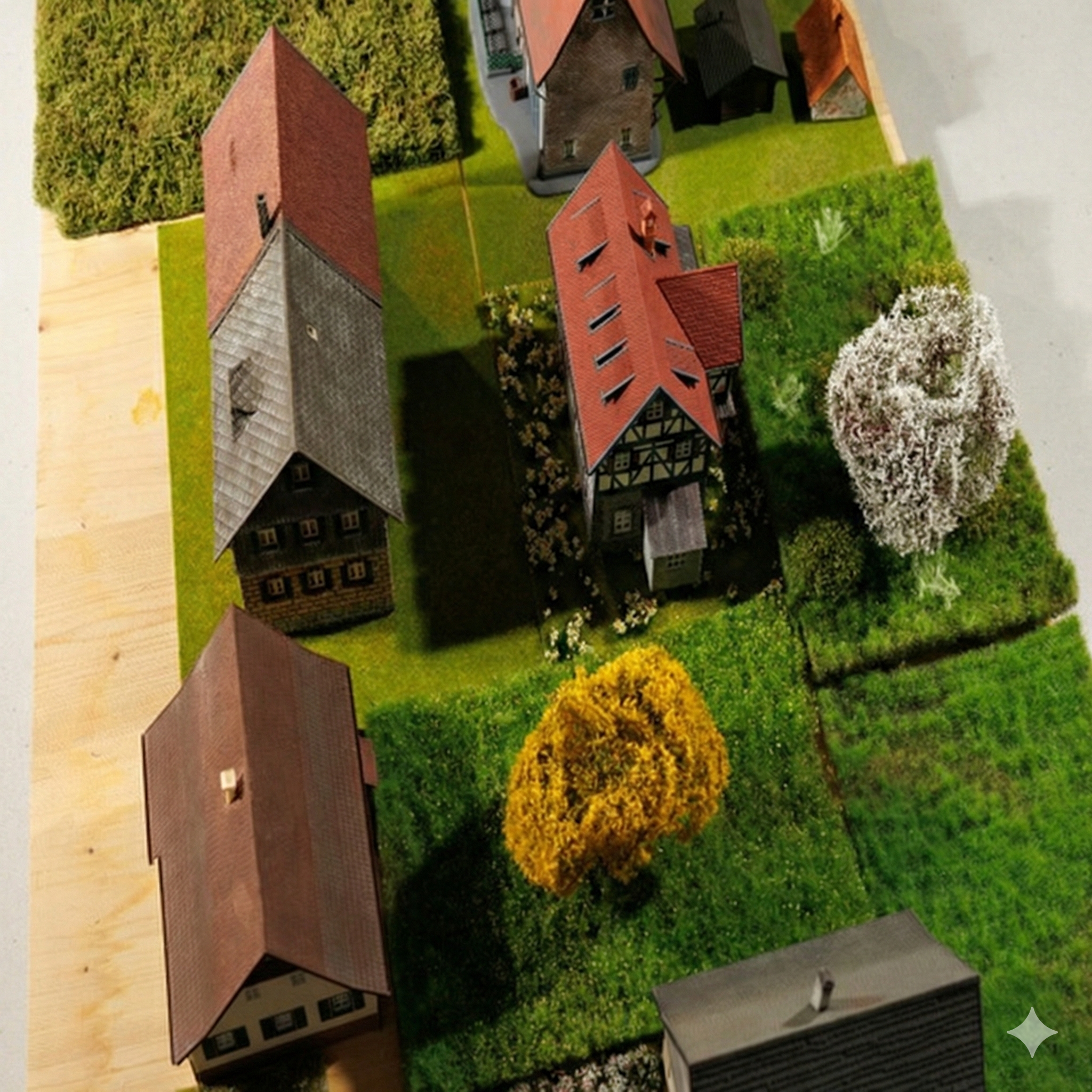} &
        \includegraphics[width=0.19\linewidth]{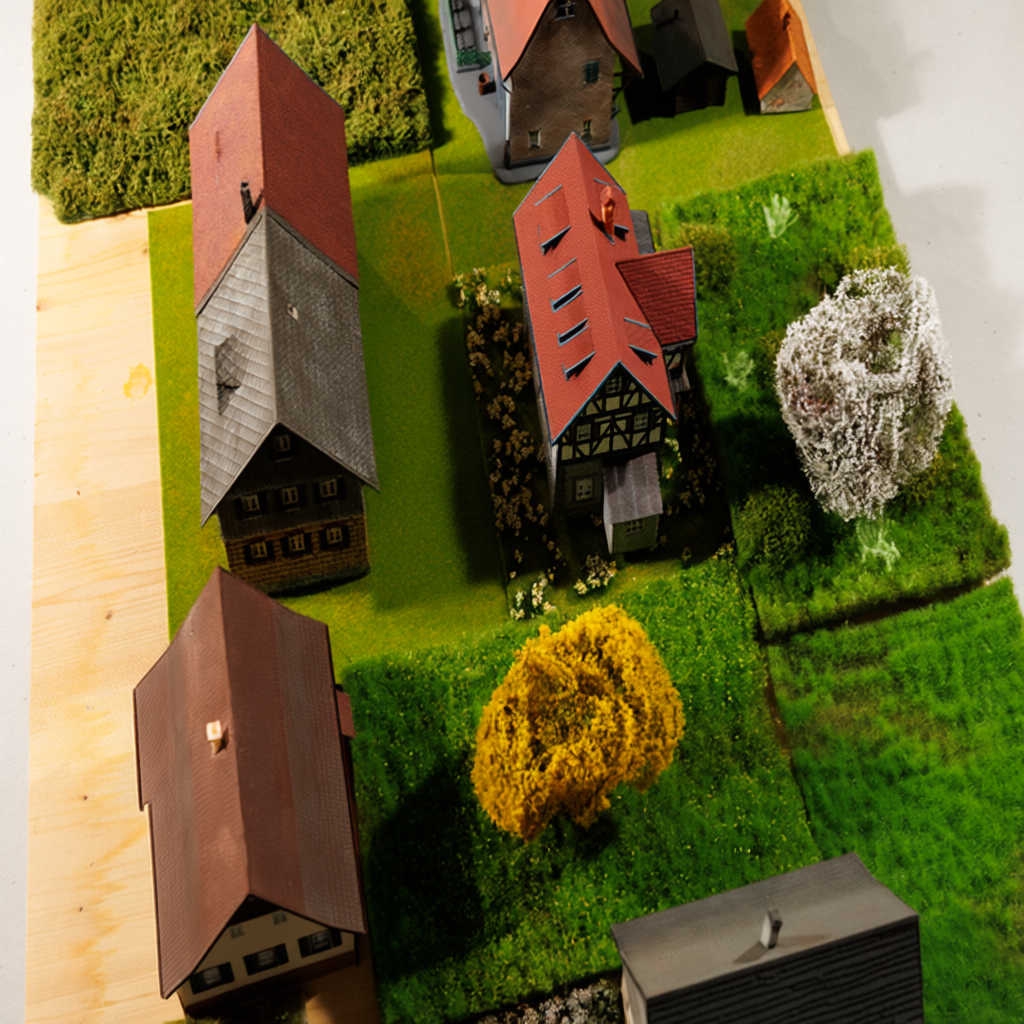} \\
        \includegraphics[width=0.19\linewidth]{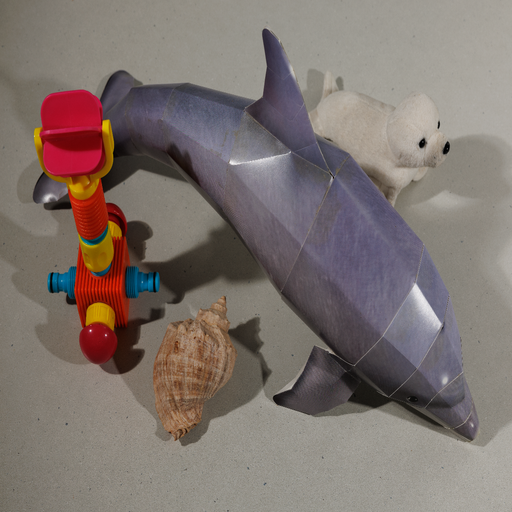} &
        \includegraphics[width=0.19\linewidth]{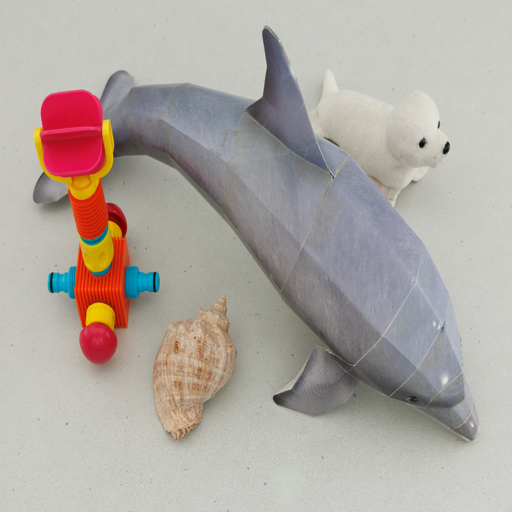} &
        \includegraphics[width=0.19\linewidth]{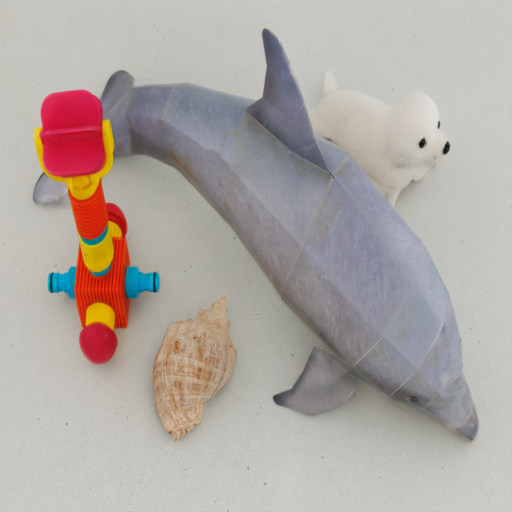} &
        \includegraphics[width=0.19\linewidth]{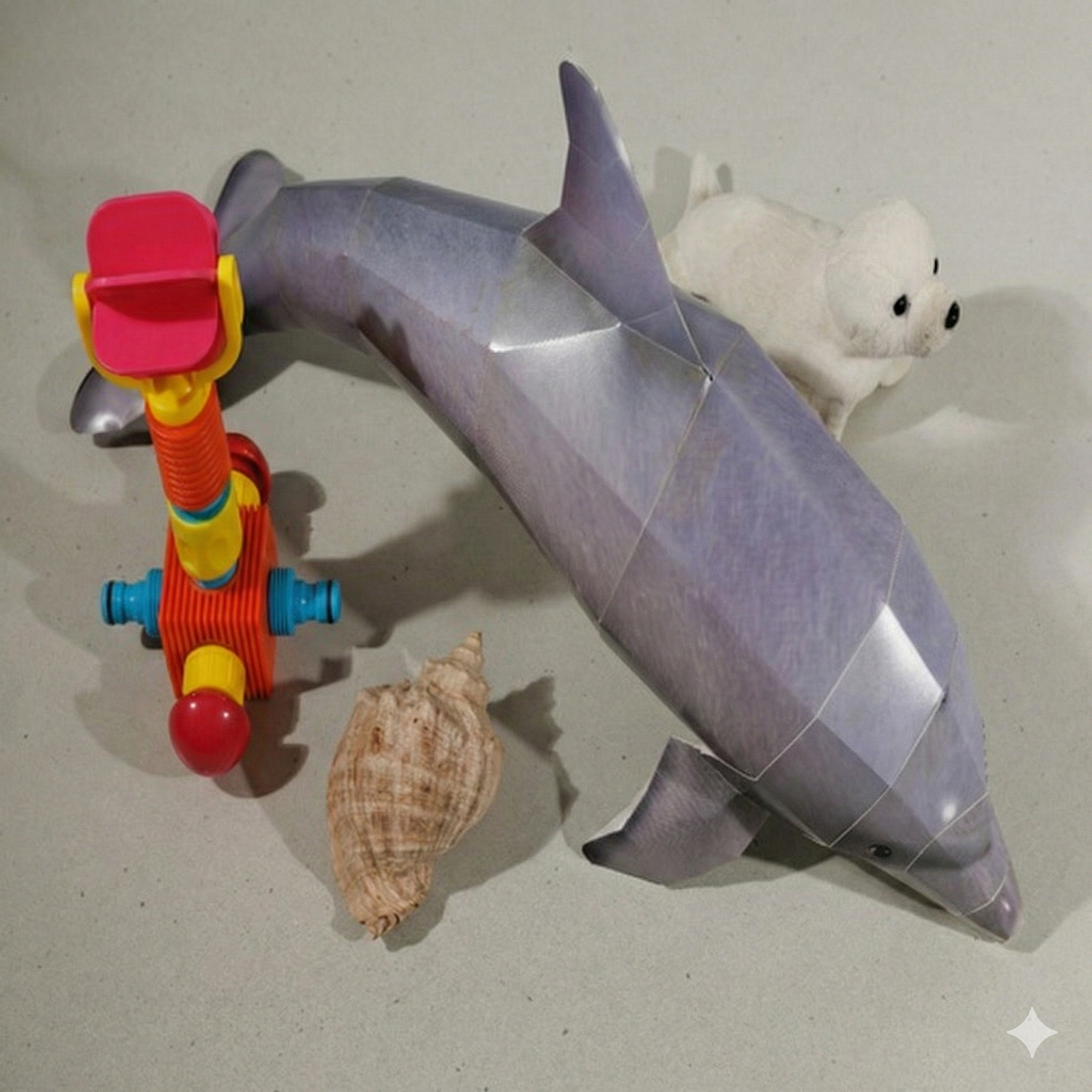} &
        \includegraphics[width=0.19\linewidth]{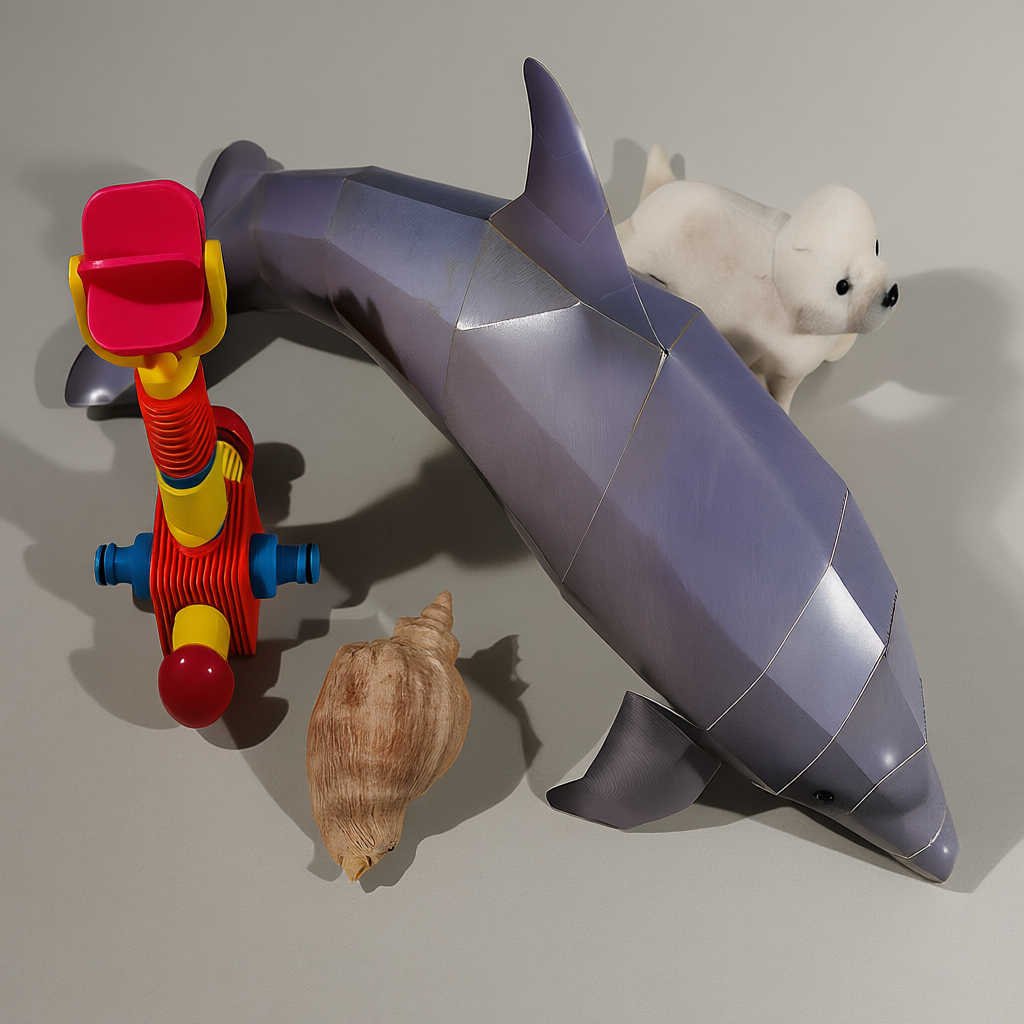} \\
        \scriptsize Input & \scriptsize GT & \scriptsize PhaSR (Ours) & \scriptsize Nano Banana & \scriptsize Qwen
    \end{tabular}
    \vspace{-0.5em}
    \captionof{figure}{\textbf{Comparison between PhaSR and Multimodal LLMs.} We select two sample images from Ambient6K, PhaSR (29.41dB, 0.2s) significantly outperforms Qwen-image-edit (15.43dB, 25m) and Nano Banana (16.86dB, 30s) in both fidelity and speed (both prompted with "Remove shadows from the image").}
    \label{fig:multimodal}
    \vspace{-1em}
\end{figure}

\noindent\textbf{Comparison with Multimodal LLMs.}
As shown in Fig.~\ref{fig:multimodal}, despite their general image editing 
capabilities, MLLMs fall substantially short on shadow removal. When prompted 
with ``Remove shadows from the image'', Qwen-image-edit and Nano Banana 
achieve only 15.43 dB and 16.86 dB on Ambient6K, respectively, while PhaSR 
reaches 29.41 dB --- a margin of over 12 dB. Furthermore, PhaSR completes 
inference in 0.2 seconds, compared to 30 seconds and 25 minutes for the MLLM 
counterparts. These results suggest that general-purpose MLLMs lack the 
task-specific physical priors necessary for precise shadow removal, whereas 
PhaSR's geometric-semantic alignment enables both superior fidelity and 
practical efficiency.

\section{Additional Feature Map Comparison}
\label{sec:feature_comparison}

\begin{figure*}[t!]
\centering
\includegraphics[width=1.0\linewidth]{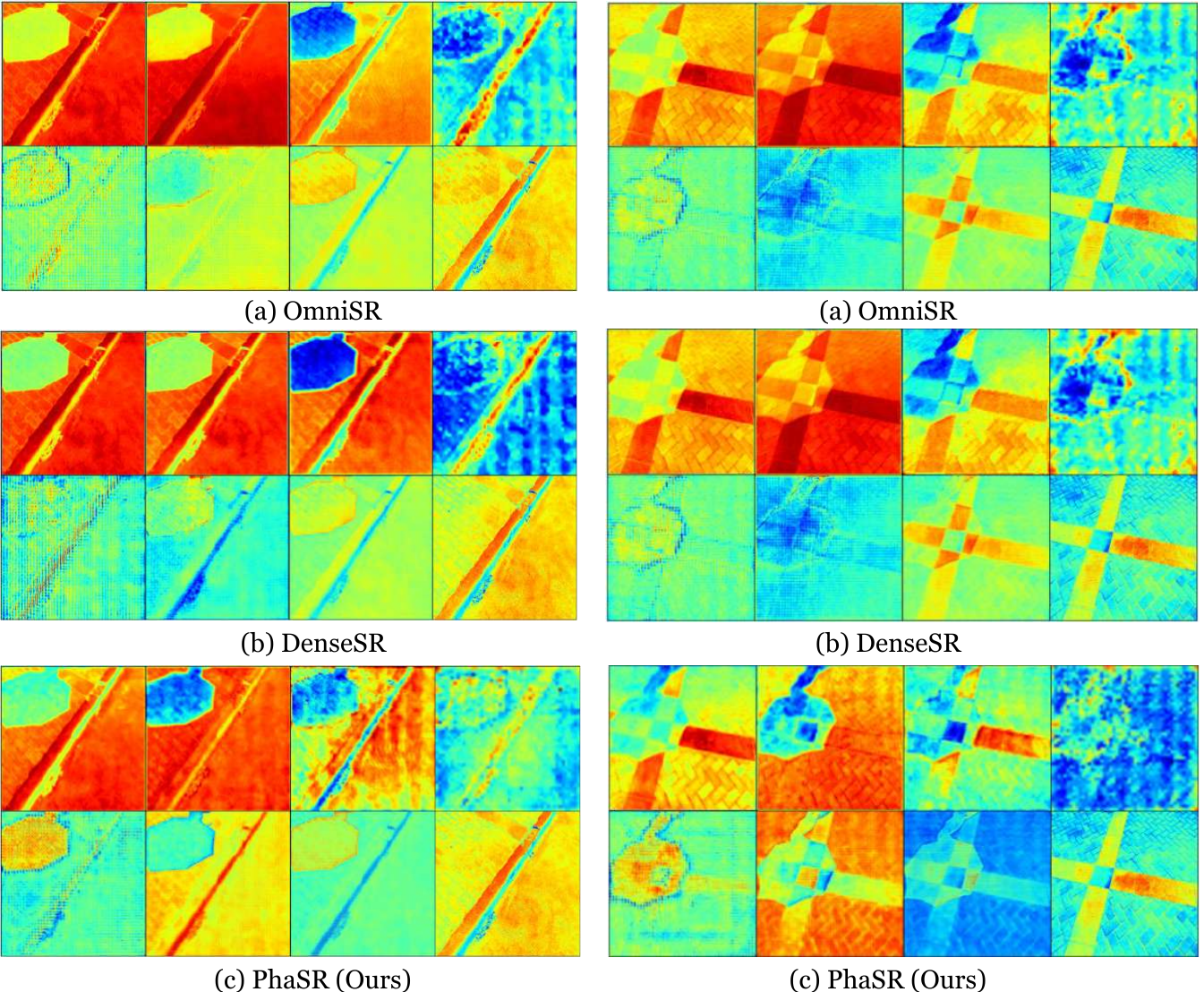}
\caption{\textbf{Intermediate feature map visualization on ISTD+~\cite{le2019shadow}.} Our method shows stronger shadow localization in bottleneck features and cleaner decoder activations compared to OmniSR \cite{omnisr} and DenseSR \cite{densesr}, validating the effectiveness of physically aligned prior propagation.}
\label{fig:featuremap_supplementary} 
\vspace{-3mm}
\end{figure*}

Figure~\ref{fig:featuremap_supplementary} visualizes intermediate feature maps from the encoder and decoder stages across different methods. Compared to OmniSR \cite{omnisr} and DenseSR \cite{densesr}, PhaSR's feature maps suggest several potential advantages:
\begin{itemize}
\item \textbf{Shadow localization:} The bottleneck features show more focused activations in shadow regions, even under complex ambient lighting.
\item \textbf{Prior propagation:} Geometric and semantic information appears well-preserved through skip connections via GSRA.
\item \textbf{Decoder activations:} The decoder shows progressive refinement with reduced high-frequency noise.
\end{itemize}

These visualizations provide qualitative evidence that the proposed physically aligned design may enable more coherent multi-scale feature learning for shadow removal.

\begin{figure}[t]
\centering
\includegraphics[width=1.0\linewidth]{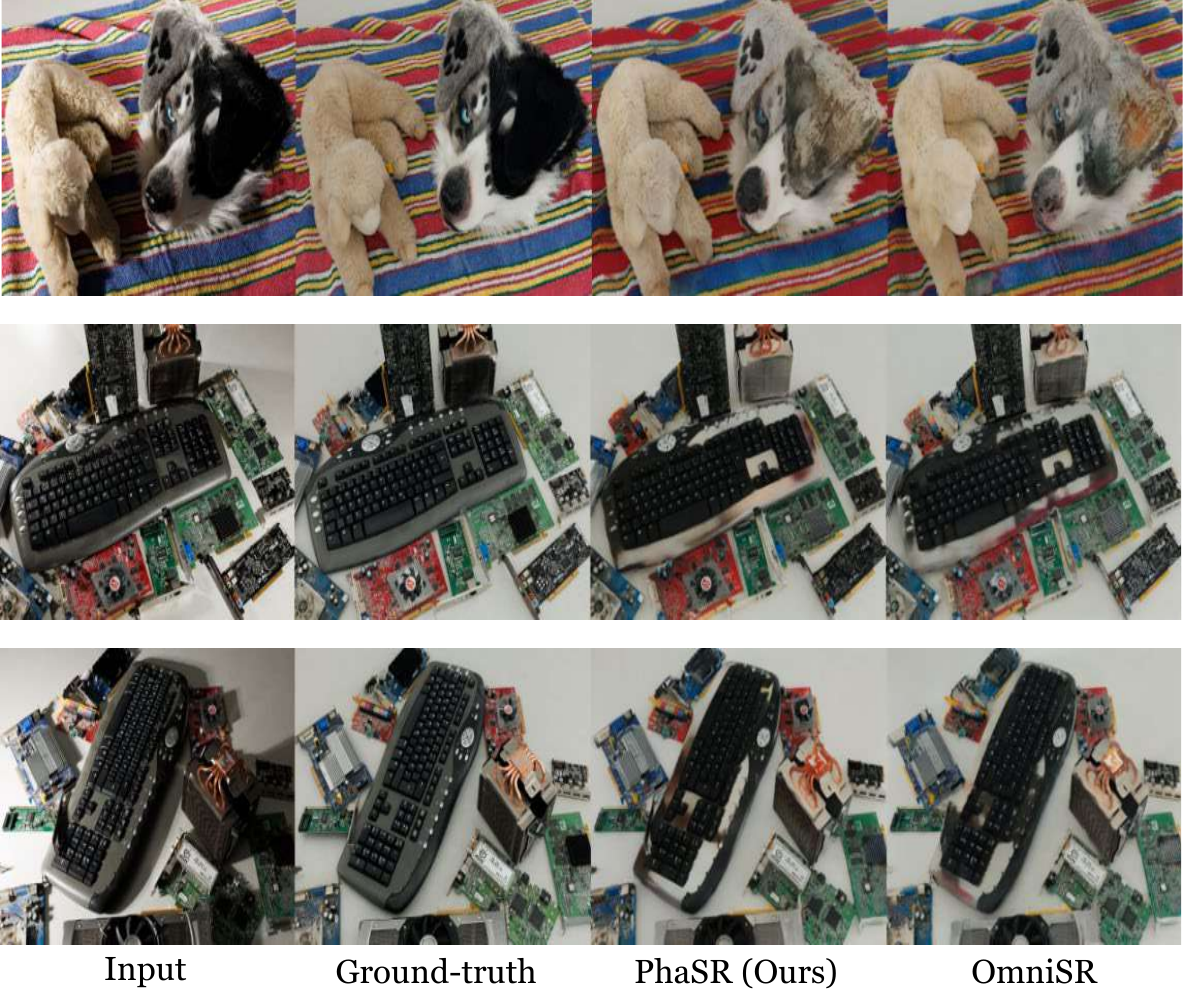}
\caption{\textbf{Failure cases on Ambient6K~\cite{ambient6k}.} Both PhaSR and existing methods struggle with shadows on intrinsically dark objects (top) or specular/metallic surfaces (bottom).}
\label{fig:failurecase} 
\vspace{-5mm}
\end{figure}

\section{Failure Case Study}

Despite competitive performance across datasets, certain scenarios remain challenging for current shadow removal methods. As shown in Figure~\ref{fig:failurecase}, both PhaSR and state-of-the-art approaches like OmniSR~\cite{omnisr} encounter difficulties in two cases:

\textbf{Dark intrinsic materials.} Shadows on low-reflectance objects (e.g., black surfaces) create ambiguity between shadow-induced darkness and intrinsic material properties. Without additional cues like polarization, methods struggle to distinguish these cases, leading to under-correction or over-brightening.

\textbf{Specular surfaces.} Metallic and specular materials violate Lambertian assumptions underlying most shadow removal methods. View-dependent highlights and non-linear light transport cause color artifacts and inconsistent restoration when shadows interact with such surfaces.

These challenges suggest future directions including material-aware priors and non-Lambertian reflectance modeling for ambient light normalization.

\section{Network Architecture Details}
\label{sec:architecture}

We provide the complete architecture specification of PhaSR in Table~\ref{tab:architecture}. The network consists of six main stages: physically aligned normalization, prior extraction, multi-scale encoder with prior fusion, bottleneck, hierarchical decoder with GSRA, and output generation.

\begin{table*}[t]
\centering
\caption{\textbf{Architecture of PhaSR.} The model takes a $H \times W$ input image and processes it through PAN normalization, multi-scale Transformer encoder-decoder with DINO-V2 semantic priors and depth-derived geometric priors.}
\label{tab:architecture}
\resizebox{0.8\textwidth}{!}{
\begin{tabular}{l|c|c|c}
\hline
\textbf{Block Name} & \textbf{Output Size} & \textbf{Operation} & \textbf{Stage} \\
\hline
\hline
\multicolumn{4}{c}{\textit{Stage 1: Physically Aligned Normalization (PAN)}} \\
\hline
Global Estimation & $H \times W \times 3$ & $\mathbf{I}_{\mathrm{norm}} = \mathbf{I} / (\mathbb{E}[\mathbf{I}] + \epsilon)$ & Eq. 2 \\
Local Normalization & $H \times W \times 3$ & $\mathbf{G}(x) = \mathbb{E}[\mathbf{I}] / (\mathbb{E}_{\Omega(x)}[\mathbf{I}] + \epsilon)$ & Eq. 3 \\
Log-domain Decomposition & $H \times W \times 3$ & $\log \hat{\mathbf{S}}, \log \hat{\mathbf{R}}$ separation & Eq. 4-5 \\
Recombination & $H \times W \times 3$ & $\hat{\mathbf{I}} = \mathrm{clamp}(\hat{\mathbf{R}} \otimes \hat{\mathbf{S}}, 0, 1)$ & Eq. 5 \\
\hline
\hline
\multicolumn{4}{c}{\textit{Stage 2: Prior Extraction}} \\
\hline
\textbf{Semantic Prior (DINO-V2)} & & & \\
DINO Scale 0 & $H/1 \times W/1 \times 1024$ & Frozen pretrained features & $\mathbf{F}_{\mathrm{D}}^{(0)}$ \\
DINO Scale 1 & $H/2 \times W/2 \times 1024$ & Frozen pretrained features & $\mathbf{F}_{\mathrm{D}}^{(1)}$ \\
DINO Scale 2 & $H/4 \times W/4 \times 1024$ & Frozen pretrained features & $\mathbf{F}_{\mathrm{D}}^{(2)}$ \\
DINO Scale 3 & $H/8 \times W/8 \times 1024$ & Frozen pretrained features & $\mathbf{F}_{\mathrm{D}}^{(3)}$ \\
\hline
\textbf{Geometric Prior} & & & \\
Depth Extraction & $H \times W \times 1$ & DepthAnything-V2 & $\mathbf{D}$ \\
Normal Computation & $H \times W \times 3$ & Gradient-based $\nabla \mathbf{D}$ & $\mathbf{N}$ \\
\hline
\hline
\multicolumn{4}{c}{\textit{Stage 3: Multi-Scale Encoder with Prior Fusion}} \\
\hline
Input Projection & $H \times W \times C$ & Conv $4 \rightarrow C$, $C=32$ & $\mathbf{y}_0$ \\
\hline
\textbf{Encoder Level 0} ($H \times W$) & & & \\
DINO Projection & $H \times W \times C$ & Conv$_{1 \times 1}$: $1024 \rightarrow C$ & $\alpha_0$ \\
TEB (CA+DWT) $\times N_1$ & $H \times W \times C$ & $N_1=2$ layers & $\mathbf{c}_0$ \\
Downsample & $H/2 \times W/2 \times 2C$ & Conv$_{4 \times 4}$, stride=2 & -- \\
\hline
\textbf{Encoder Level 1} ($H/2 \times W/2$) & & & \\
DINO Projection & $H/2 \times W/2 \times 2C$ & Conv$_{1 \times 1}$: $1024 \rightarrow 2C$ & $\alpha_1$ \\
TEB (CA+DWT) $\times N_2$ & $H/2 \times W/2 \times 2C$ & $N_2=2$ layers & $\mathbf{c}_1$ \\
Downsample & $H/4 \times W/4 \times 4C$ & Conv$_{4 \times 4}$, stride=2 & -- \\
\hline
\textbf{Encoder Level 2} ($H/4 \times W/4$) & & & \\
DINO Projection & $H/4 \times W/4 \times 4C$ & Conv$_{1 \times 1}$: $1024 \rightarrow 4C$ & $\alpha_2$ \\
TEB (GSRA) $\times N_3$ & $H/4 \times W/4 \times 4C$ & $N_3=2$ layers & $\mathbf{c}_2$ \\
Downsample & $H/8 \times W/8 \times 8C$ & Conv$_{4 \times 4}$, stride=2 & -- \\
\hline
\hline
\multicolumn{4}{c}{\textit{Stage 4: Bottleneck ($H/8 \times W/8$)}} \\
\hline
Multi-Scale DINO Fusion & $H/8 \times W/8 \times 8C$ & Concat + Conv$_{1 \times 1}$: $4096 \rightarrow 8C$ & $\mathbf{F}_{\mathrm{cat}}$ \\
DINO Projection Level 3 & $H/8 \times W/8 \times 8C$ & Conv$_{1 \times 1}$: $1024 \rightarrow 8C$ & $\alpha_3$ \\
PATB (GSRA) $\times N_4$ & $H/8 \times W/8 \times 16C$ & $N_4=2$ layers, concat input & $\mathbf{c}_3$ \\
\hline
\hline
\multicolumn{4}{c}{\textit{Stage 5: Hierarchical Decoder with GSRA}} \\
\hline
\textbf{Decoder Level 2} ($H/4 \times W/4$) & & & \\
Upsample & $H/4 \times W/4 \times 4C$ & ConvTranspose$_{2 \times 2}$, stride=2 & -- \\
Skip Connection & $H/4 \times W/4 \times 8C$ & Concat with $\mathbf{c}_2$ & $\mathbf{u}_2$ \\
GSRA (Sec. 3.2) & $H/4 \times W/4 \times 8C$ & Geometric-Semantic Rectification & Eq. 6-10 \\
TDB (CA+DWT) $\times N_5$ & $H/4 \times W/4 \times 8C$ & $N_5=2$ layers & $\mathbf{c}_2'$ \\
\hline
\textbf{Decoder Level 1} ($H/2 \times W/2$) & & & \\
Upsample & $H/2 \times W/2 \times 2C$ & ConvTranspose$_{2 \times 2}$, stride=2 & -- \\
Skip Connection & $H/2 \times W/2 \times 4C$ & Concat with $\mathbf{c}_1$ & $\mathbf{u}_1$ \\
GSRA (Sec. 3.2) & $H/2 \times W/2 \times 4C$ & Geometric-Semantic Rectification & Eq. 6-10 \\
TDB (CA+DWT) $\times N_6$ & $H/2 \times W/2 \times 4C$ & $N_6=2$ layers & $\mathbf{c}_1'$ \\
\hline
\textbf{Decoder Level 0} ($H \times W$) & & & \\
Upsample & $H \times W \times C$ & ConvTranspose$_{2 \times 2}$, stride=2 & -- \\
Skip Connection & $H \times W \times 2C$ & Concat with $\mathbf{c}_0$ & $\mathbf{u}_0$ \\
GSRA (Sec. 3.2) & $H \times W \times 2C$ & Geometric-Semantic Rectification & Eq. 6-10 \\
TDB (CA+DWT) $\times N_7$ & $H \times W \times 2C$ & $N_7=2$ layers & $\mathbf{c}_0'$ \\
\hline
\hline
\multicolumn{4}{c}{\textit{Stage 6: Output Generation}} \\
\hline
Output Projection & $H \times W \times 3$ & Conv$_{3 \times 3}$: $2C \rightarrow 3$ & -- \\
Residual Connection & $H \times W \times 3$ & $\hat{\mathbf{I}} = \mathrm{OutProj}(\mathbf{c}_0') + \mathbf{I}$ & Final \\
\hline
\end{tabular}
}
\end{table*}

{
    \small
    \bibliographystyle{ieeenat_fullname}
    \bibliography{main}
}
\end{document}